Preprint, under review.

# KEPT: Knowledge-Enhanced Prediction of Trajectories from Consecutive Driving Frames with Vision-Language Models

Yujin Wang[a], Tianyi Wang[b], Quanfeng Liu[a], Wenxian Fan[a], Junfeng Jiao[c], Christian Claudel[b], Yunbing Yan[d], Bingzhao Gao[a,*], Jianqiang Wang[e], and Hong Chen[f]

[a] *School of Automotive Studies, Tongji University, Shanghai, 201804, China*

[b] *Department of Civil, Architectural, and Environmental Engineering, The University of Texas at Austin, Austin, Texas, 78712, USA*

[c] *School of Architecture, The University of Texas at Austin, Austin, Texas, 78712, USA*

[d] *School of Automotive and Traffic Engineering, Wuhan University of Science and Technology, Wuhan, 430081, China*

[e] *School of Vehicle and Mobility, Tsinghua University, Beijing, 100084, China*

[f] *College of Electronic and Information Engineering, Tongji University, Shanghai, 201804, China*

* *Corresponding author.*

*E-mail address:* gaobz@tongji.edu.cn

**Abstract**

Accurate short-horizon trajectory prediction is crucial for safe and reliable autonomous driving. However, existing vision-language models (VLMs) often fail to accurately understand driving scenes and generate trustworthy trajectories. To address this challenge, this paper introduces KEPT, a knowledge-enhanced VLM framework that predicts ego trajectories directly from consecutive front-view driving frames. KEPT integrates a temporal frequency–spatial fusion (TFSF) video encoder, which is trained via self-supervised learning with hard-negative mining, with a k-means & HNSW retrieval-augmented generation (RAG) pipeline. Retrieved prior knowledge is added into chain-of-thought (CoT) prompts with explicit planning constraints, while a triple-stage fine-tuning paradigm aligns the VLM backbone to enhance spatial perception and trajectory prediction capabilities. Evaluated on nuScenes dataset, KEPT achieves the best open-loop performance compared with baseline methods. Ablation studies on fine-tuning stages, Top-K value of RAG, different retrieval strategies, vision encoders, and VLM backbones are conducted to demonstrate the effectiveness of KEPT. These results indicate that KEPT offers a promising, data-efficient way toward trustworthy trajectory prediction in autonomous driving.

**Keywords**

Autonomous Driving · Trajectory Prediction · Vision-Language Model · Retrieval-Augmented Generation · Chain-of-Thought Prompt

# Nomenclature

| **Symbol** | **Meaning (units)** |
|---|---|
| **Sets, indices, and dimensions** | |
| $H, W, P$ | Image height, width, patch size (pixels) |
| $T$ | Number of frames per clip (here $T = 7$) |
| $N_p$ | Number of patches per frame ($N_p = \frac{H}{P}\frac{W}{P}$) |
| $\mathbb{R}, \mathbb{C}$ | Real and complex domains |
| **Images, patches, and transforms** | |
| $I_{RGB} \in \mathbb{R}^{3\times H\times W}$ | Input RGB image (normalized intensity) |
| $I_g \in \mathbb{R}^{1\times H\times W}$ | Grayscale image |
| $X_{\text{patch}}^{(i)} \in \mathbb{R}^{P\times P}$ | $i$-th grayscale patch |
| $\mathscr{F}\{\cdot\}$ | 2D FFT operator |
| $F^{(i)} = \mathscr{F}\{X_{\text{patch}}^{(i)}\} \in \mathbb{C}^{P\times P}$ | Patch spectrum |
| $A^{(i)} = \lvert F^{(i)}\rvert \in \mathbb{R}^{P\times P}$ | Amplitude spectrum |
| $\text{GAP}(\cdot)$ | Global average pooling |
| $a^{(i)} = \text{GAP}(A^{(i)}) \in \mathbb{R}$ | Amplitude statistic of patch $i$ |
| $\mathbf{w}_{\text{freq}} \in [0,1]^{N_p}$ | Frequency-domain attention weights |
| $X_f^{(i)} = w_{\text{freq}}^{(i)} X_{\text{patch}}^{(i)}$ | Re-weighted patch |
| $X_f \in \mathbb{R}^{1\times H\times W}$ | Frequency-attended image |

| Symbol | Meaning (units) |
|---|---|
| $X_{\text{spa}}^{(\ell)}$ | Swin-Tiny stage-$\ell$ spatial feature map |
| $X_s$ | Upsampled / Concatenated spatial representation |
| $X_{fs} = [X_f; X_s]$ | Channel concatenation of $X_f$ and $X_s$ |
| $X_{\text{feat}} \in \mathbb{R}^{C' \times H \times W}$ | Fused representation after $1 \times 1$ conv, BN, ReLU |
| **Temporal encoding** | |
| $\mathbf{h}_t = \text{GAP}(X_{\text{feat},t}) \in \mathbb{R}^{C'}$ | Frame-$t$ feature |
| $\mathbf{H}_{\text{seq}} = [\mathbf{h}_1; \ldots; \mathbf{h}_T] \in \mathbb{R}^{T \times C'}$ | Temporal feature matrix |
| $\mathbf{H}_{\text{in}}, \mathbf{H}_{\text{out}}$ | Transformer input / output sequences |
| $h_{\text{TFSF}} \in \mathbb{R}^{C'}$ | CLS-token feature used as final vector |
| **Self-supervised TFSF training (InfoNCE)** | |
| $\mathscr{E}_\theta$ | TFSF backbone with parameters $\theta$ |
| $\mathscr{G}_\phi$ | Projection head with parameters $\phi$ |
| $\mathbf{z} = \dfrac{\mathscr{G}_\phi(\mathbf{h})}{\lVert \mathscr{G}_\phi(\mathbf{h}) \rVert_2} \in \mathbb{R}^m$ | $\ell_2$-normalized projection on unit hypersphere |
| $\text{sim}(\mathbf{u}, \mathbf{v}) = \mathbf{u}^\top \mathbf{v}$ | Inner product (cosine similarity numerator) |
| $\tau_{\text{temp}}$ | Temperature for InfoNCE |
| $\mathscr{N}_i^B$, $\mathscr{Q}$ | In-batch negatives; memory queue (capacity $M$) |
| $\text{TopK}(\cdot, K)$ | Hard-negative mining (keep top-$K$) |
| $\mathscr{L}_i$ | InfoNCE loss of sample $i$; $L = (1/B) \sum_i \mathscr{L}_i$ |
| $B$, $E$, $M$, $K$ | Batch size; #epochs; queue capacity; #hard negatives |
| **Embedding & retrieval** | |
| $\mathbf{e}_i \in \mathbb{R}^{128}$ | Normalized embedding of clip $i$ |
| $n_c$, $\mathbf{c}_j$ | #Clusters; k-means centroid of cluster $j$ |
| $d_{\cos}(\mathbf{x}, \mathbf{y}) = 1 - \dfrac{\mathbf{x}^\top \mathbf{y}}{\lVert \mathbf{x} \rVert_2 \lVert \mathbf{y} \rVert_2}$ | Cosine distance |
| $\text{HNSW}(M,\ \text{ef}_{\text{construction}})$ | HNSW index (connectivity $M$, build $ef$) |
| $k$ | #Nearest neighbors for retrieval |
| **Triple-stage fine-tuning (Sec. 3.4)** | |
| *Stage A: Spatial perception (VQA multi-task)* | |

| **Symbol** | **Meaning (units)** |
|---|---|
| $\mathscr{L}_{\mathrm{A}}$ | Stage-A multi-task loss (classification + size + distance) |
| $K$ | Number of semantic categories for classification |
| $\eta_{k,t},\ \pi_{k,t}$ | One-hot GT label and predicted probability for class $k$ in sample $t$ |
| $s_{m,t},\ \hat{s}_{m,t}$ | Predicted / Reference object size along dimension $m \in \{1,2,3\}$ |
| $d_t,\ \hat{d}_t$ | Predicted / reference distance to ego vehicle |
| $\alpha_t,\ \beta_t,\ \gamma_t$ | Binary indicators activating cls. / size / dist. terms for sample $t$ |
| *Stage B: Surround-view trajectory regression* | |
| $\mathscr{L}_B$ | Stage-B trajectory regression objective (coords + velocity; with smoothness / curvature regularization) |
| $Q$ | Number of prediction horizons (here $Q = 3$, at 1s / 2s / 3s) |
| $q,\ r$ | Horizon index $q \in \{1,\ldots,Q\}$; sample index $r \in \{1,\ldots,B\}$ |
| $x_{q,r},\ y_{q,r},\ v_{q,r}$ | Predicted ego-centric $x/y$ displacements and velocity at horizon $q$ for sample $r$ |
| $\tilde{x}_{q,r},\ \tilde{y}_{q,r},\ \tilde{v}_{q,r}$ | Ground-truth counterparts of $x_{q,r}, y_{q,r}, v_{q,r}$ |
| *Stage C: Front-view end-to-end trajectory prediction* | |
| $\mathscr{L}_C$ | Stage-C loss, reusing Stage-B objective ($\mathscr{L}_C = \mathscr{L}_B$) |
| **Kinematics and evaluation** | |
| $(x,y)$ | Ego-centric coordinates (m) |
| $v$ | Velocity (km/h) |
| $\hat{\tau} = \{(\hat{x}_i,\hat{y}_i)\}_{i=1}^{6},\ \tau = \{(x_i,y_i)\}_{i=1}^{6}$ | Predicted / Ground-truth trajectory at 2Hz over 3s (m) |
| $l_i = \sqrt{(x_i-\hat{x}_i)^2+(y_i-\hat{y}_i)^2}$ | Per-step Euclidean error (m) |
| $\bar{l}_{2,i}$ | Mean $l_2$ at step $i$ (m) |
| $L_{2,k}^{\mathrm{UniAD}},\ L_{2,k}^{\mathrm{VAD}}$ | L2 error metrics (UniAD at step $2k$; VAD cumulative to $k$) (m) |

| **Symbol** | **Meaning (units)** |
|---|---|
| $C_i$, $\bar{C}_i$ | Collision indicator (0 / 1) and its average |
| $C_k^{\text{UniAD}}$, $C_k^{\text{VAD}}$ | Collision metrics (UniAD at $2k$; VAD cumulative) |

# 1 Introduction

Accurate prediction of vehicle trajectories is a fundamental capability of autonomous driving systems, crucial for safe and efficient navigation in dynamic traffic environments Chai et al. (2020); Ngiam et al. (2022); Wang and Wang (2024). Recently, end-to-end autonomous driving methods have emerged as a promising paradigm by directly taking sensor data as input for perception and output planning results with one holistic model Liu et al. (2021a); Chib and Singh (2023); Chen et al. (2024b). Through extensive data training, end-to-end approaches have demonstrated impressive planning capabilities, providing a streamlined and competitive alternative to traditional modular pipelines Casas et al. (2021); Chen and Krähenbühl (2022); Yang et al. (2025a).

While end-to-end approaches have been widely adopted and constantly making breakthroughs on challenging benchmarks, they solely rely on fixed-format inputs, which restricts the agent's ability to comprehend multimodal information and interact with the environment and human users Shao et al. (2023a); Yu et al. (2025); Wang et al. (2025b). Moreover, certain methods skip explicit scene understanding and directly predict driving commands from sensor data Hu et al. (2022b); Weng et al. (2024); Huang et al. (2024b), which sacrifices interpretability and introduces challenges in optimization and safety validation. Along this line of work, UniAD Hu et al. (2023) introduced a query-based design that integrates perception and prediction components, enabling an end-to-end planning scheme. VAD Jiang et al. (2023a) improved computational efficiency and interpretability by adopting a vectorized scene representation, replacing the dense rasterized representations from UniAD. Despite the remarkable progress of end-to-end models in autonomous driving, such approaches inherently struggle in long-tail scenarios, where prediction errors compound, severely degrading downstream planning performance Chitta et al. (2021); Wu et al. (2022).

Concurrently, large language models (LLMs) have demonstrated impressive capabilities in language comprehension and reasoning, showing potential to solve this problem Wei et al. (2022); Hegde et al. (2025). Going beyond text-based prompting, multimodal large language models (MLLMs) integrate image and video inputs to the LLM, enabling tasks such as visual question answer (VQA) and dense captioning Li et al. (2023a); Liu et al. (2023). Nevertheless, existing MLLM datasets for autonomous driving remain limited in scale, typically containing fewer than one million VQAs Wu et al. (2024). Consequently, without careful model and training scheme designs, these MLLMs exhibit poor performance in reasoning and planning tasks due to lacking scene understanding and grounding capability Ma et al. (2024b).

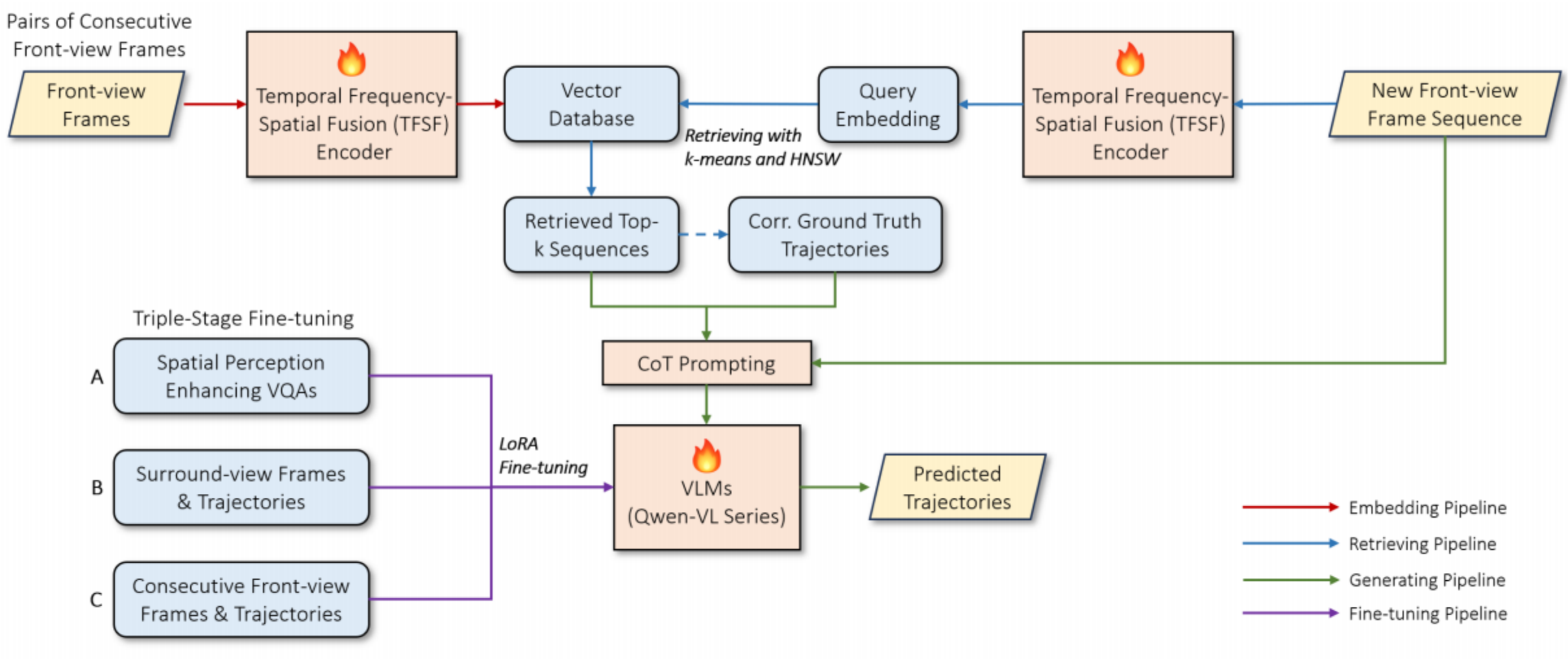

**Fig. 1** The overview of our KEPT (Knowledge-Enhanced Prediction of Trajectories) method. The method consists of four pipelines: **(1) The embedding pipeline** encodes front-view frame sequences into a vector database via a self-designed and trained temporal frequency-spatial fusion (TFSF) encoder. **(2) The retrieving pipeline** retrieves the most similar scene from the database via k-means clustering and HNSW searching. **(3) The generating pipeline** guides vision language models (VLMs) in predicting trajectories in the future 3 seconds according to the new frame sequence and retrieved information via a chain-of-thought (CoT) prompting strategy. **(4) The fine-tuning pipeline** involves a triple-stage fine-tuning paradigm, which aims to equip VLMs with spatial grounding, motion feasibility, and temporally conditioned planning capabilities, enabling trajectory prediction from raw perceptual inputs. The fire icon denotes components that are trained or fine-tuned in this work; others are frozen or non-parametric. In our pipeline, the TFSF encoder is trained with self-supervision, the retrieval stack is non-parametric, and the VLM's language head is fine-tuned with LoRA across three stages.

In this work, we propose **KEPT**, a knowledge-enhanced trajectory prediction framework, introducing a novel approach to assist an MLLM in predicting future trajectories according to consecutive driving frames, as depicted in Figure 1. The main research contributions of this work are outlined as follows:

- **Temporal Frequency–Spatial Fusion (TFSF) Encoder for Consecutive Driving Frames:** We build a video encoder that mixes frequency cues from a fast Fourier transform (FFT)-based attention module with the multi-scale spatial features of a Swin-Tiny backbone. A short temporal transformer then processes seven frames sampled at 2 Hz. Together, these components produce clip-level embeddings that capture the motion patterns most relevant for planning.
- **Self-Supervised Representation Learning with Hard-Negative Mining and Memory Queue:** We train the TFSF module with a contrastive InfoNCE loss. The batch provides most of the negatives, and we keep an additional fixed-size queue to supply more varied ones. We also use a simple Top-K strategy to choose the harder negatives. The encoder outputs unit-norm embeddings, which are then used directly for retrieval.
- **Scalable Embedding–Retrieval Pipeline for Trajectory Priors:** We build a vector database of scene embeddings and run a two-step approximate nearest-neighbor search. The first step uses k-means clustering to narrow down the candidates, and the second applies HNSW with cosine distance to find the closest matches. From these results, we retrieve the Top-K similar scenes together with their ground-truth trajectories, which serve as useful priors for the planner.
- **Triple-Stage Fine-Tuning of VLMs Bridging Perception to Planning:** (i) We improve spatial perception by training with VQAs, covering attributes such as object class, size, and distance, and we apply LoRA to keep the adaptation lightweight. (ii) Then we train it to predict trajectories using the six surround-view images and a few basic kinematics. The loss mainly keeps the motion smooth and prevents unsafe bends or sudden shifts. (iii) In the last stage, the model learns to predict the whole trajectory from consecutive front-view frames. We keep the same regression loss, but here the goal is to help the language head pick up short-term temporal structure.

- **Chain-of-Thought (CoT) Prompting with Exemplar Comparison and Planning Constraints:** We provide the VLMs with several retrieved reference scenes before the target scene. The models are then guided to compare these scenes, subject to safety constraints. Finally, a trajectory in specific format is required for the models to generate.

The rest of this paper is organized as follows: Section 2 reviews related works in this field thoroughly. In Section 3, details of KEPT are introduced. In Section 4, comparative experiments, ablation studies, and case studies are designed to demonstrate the effectiveness of KEPT. Section 5 summarizes the work and discusses future research directions.

# 2 Related Work

## 2.1 Foundation Models in Autonomous Driving

Recent advancements in foundation models have demonstrated their ability to reason over complex contextual information, interpret human intent, and generate logically consistent outputs in autonomous driving Wei et al. (2022); Mao et al. (2023); Ma et al. (2024c); Zhu et al. (2025). These works investigated the capabilities of Large Language Models (LLMs) to generalize to uncommon scenarios as well as their ability to reason about driving scenes in text. “Drive as you speak” Cui et al. (2024) enriched LLMs with comprehensive environmental data from different vehicle modules, supporting safer and more context-aware driving decisions. “Driving with LLMs” Chen et al. (2024a) introduced LLMs that generated 10,000 driving scenarios for agent training. “Drive like a human” Fu et al. (2024) demonstrated LLMs’ capabilities of interacting with environments in closed-loop systems, and effectively navigating long-tail autonomous driving scenarios. Dilu Wen et al. (2024) first leveraged knowledge-driven capability in decision-making for autonomous vehicles. CCMA Zhang et al. (2025) integrated reinforcement learning for individual optimization and LLMs for regional and global coordination, significantly improving traffic flow and safety in autonomous driving scenarios.

Meanwhile, recent works have investigated the incorporation of MLLMs into end-to-end frameworks Wang et al. (2024); Xing et al. (2025); Brandstätter et al. (2025); Liao et al. (2025). Autonomous driving systems integrating multi-source information, such as vision, language, and rules, tend to exhibit greater generalization in complex or long-tail scenarios Chen et al. (2025); Wang et al. (2025c); Ma et al. (2025). VLP Pan et al. (2024) incorporated linguistic descriptions into the training process and aligned them with visual features, which significantly improved cross-domain generalization. DriveGPT4 Xu et al. (2024) utilized VLMs to predict control commands and simultaneously generate textual justifications using an iterative VQA format. To enhance embodied intelligence, LMDrive Shao et al. (2024) integrated a vision encoder with LLMs, enabling multi-modal scene understanding and natural language command execution. Omnidrive Wang et al. (2025a) introduced a 3D VLM architecture to strengthen planning and reasoning capabilities. DriveVLM Tian et al. (2025) incorporated traditional 3D perception into a multi-stage reasoning chain, combining scene description, dynamic analysis, and hierarchical planning to bridge cognitive depth with real-time control. DiMA Hegde et al. (2025) employed a knowledge distillation approach to transfer planning capabilities from MLLMs to lightweight vision-based planners via surrogate tasks.

## 2.2 End-to-End Trajectory Prediction

Early end-to-end trajectory prediction models primarily used LiDAR point clouds Casas et al. (2020); Liang et al. (2020); Wu et al. (2020). However, due to the dependence on accurate bounding box detection, these approaches often failed to generalize in the presence of occluded or unclassified objects. To solve this problem, recent studies have shifted toward vision-centric methods based on bird's-eye view (BEV) representations, which provide a unified spatial-temporal understanding of the driving scene Hu et al. (2021a); Akan and Güney (2022); Liang et al. (2022). For instance, PowerBEV Li et al. (2023b) realized semantic and instance-level trajectory prediction by simply forecasting segmentation and centripetal backward flow. TPV Huang et al. (2023) extended BEV with two additional perpendicular planes and utilized attention mechanism to aggregate the image features corresponding to each query in each TPV plane. LiDAR-camera fusion approaches have also emerged. For example, DAL Huang et al. (2024a) contained an attention-free predicting pipeline and an easy training process to achieve trade-off between speed and accuracy of 3D object detection and prediction through efficient multi-modal fusion. BEVFormer Li et al. (2025) leveraged a unified multi-modality spatial attention mechanism for integrating camera and LiDAR data with historical BEV features. Despite these advances, BEV-based methods still face limitations in capturing fine-grained 3D structures and object-level semantics.

To address these shortcomings, vision-centric 4D perception emerges as a promising alternative to effectively extend temporal occupancy prediction with camera images as input but also broaden both semantic and instance prediction beyond BEV format and predefined categories Tian et al. (2023); Tong et al. (2023); Wang et al. (2023); Wei et al. (2023). Cam4DOcc Ma et al. (2024a) was the first to achieve vision-centric 4D occupancy forecasting task, allowing simultaneous prediction of future trajectory for both general movable and static objects. Building upon this, Drive-OccWorld Yang et al. (2025b) investigated potential applications of the camera-based 4D occupancy forecasting world model by injecting action conditions and integrating this generative capability with end-to-end planners for safe driving.

### 2.3 Retrieval-Augmented Generation in Vision-Language Models

In vision-language tasks, retrieval-augmented generation (RAG) mitigates knowledge limitations by leveraging external knowledge bases, enabling models to extract insights from images while supplementing them with retrieved contextual data Shao et al. (2023b); Wang et al. (2025c). Prior studies Jiang et al. (2023b); Ram et al. (2023) on VLMs have shown that retrieval can improve contextual integration and multi-step reasoning under knowledge deficits, strengthen performance on complex question answering when paired with strong pre-trained encoders, and yield better downstream results when large external corpora are incorporated during pre-training and fine-tuning. Follow-up works Zheng et al. (2024); Hussien et al. (2025) further highlighted RAG's advantages for adaptive, multi-modal generation in data-scarce domains and its ability to tighten cross-modal associations for more faithful grounding.

For autonomous driving applications, dynamic retrieval policies that adjust what to fetch at run time have been proposed to match task requirements and latency budgets Bandyopadhyay et al. (2025). Domain knowledge can also be queried explicitly: a traffic-regulation agent retrieves applicable rules and guideline hints conditioned on ego state and scene context to inform compliance-aware planning Gao et al. (2025). RAG offers a principled way to compensate for the limited world knowledge and long-tail corner cases that arise in autonomous driving Atakishiyev et al. (2024). For instance, RAG-Driver Yuan et al. (2024) introduced a retrieval-augmented in-context learning mechanism through a curated multi-modal driving in-context instruction tuning dataset and a vector similarity-based retrieval engine specifically tailored for driving applications. VistaRAG Dai et al. (2024) positioned RAG as a safe and trustworthy layer that dynamically retrieves prior driving experience, live road-network state, and other contextual evidence to ground decisions and make the reasoning auditable. SenseRAG Luo et al. (2025) constructed an LLM-readable environmental knowledge base from multimodal sensor/V2X streams and issued proactive queries to retrieve time-critical facts, yielding measurable gains in perception and prediction under latency constraints.

In summary, recent advances show that VLMs and RAG substantially enhance grounding, contextual reasoning, and data efficiency for autonomous driving; however, most pipelines (i) retrieve generic knowledge rather than planning-specific cues, (ii) model temporal dynamics weakly for long-horizon forecasts, and (iii) offer limited safety calibration and auditability when reasoning under distribution shift. In addition, the training signal often fails to reflect metric accuracy or basic physical constraints, which makes the model behave unreliably in crowded scenes where rules and interactions matter.

To address these limitations, we introduce KEPT, a retrieval-augmented VLM planner that couples a temporal frequency-spatial fusion encoder with scene-aligned retrieval, CoT prompting and a triple-stage fine-tuning strategy that grounds reasoning in metric trajectories and collision risks. KEPT is designed to turn externally retrieved knowledge into actionable, long-horizon plans, improving both accuracy and safety.

# 3 Methodology

## 3.1 Temporal Frequency-Spatial Fusion Encoder for Consecutive Driving Frames

In this subsection, we introduce a novel TFSF encoder for consecutive driving frames. The TFSF encoder integrates frequency-domain and spatial-domain representations at multiple granularities and employs a temporal transformer for encoding temporal dependencies. The overview of the architecture is illustrated in Figure 2.

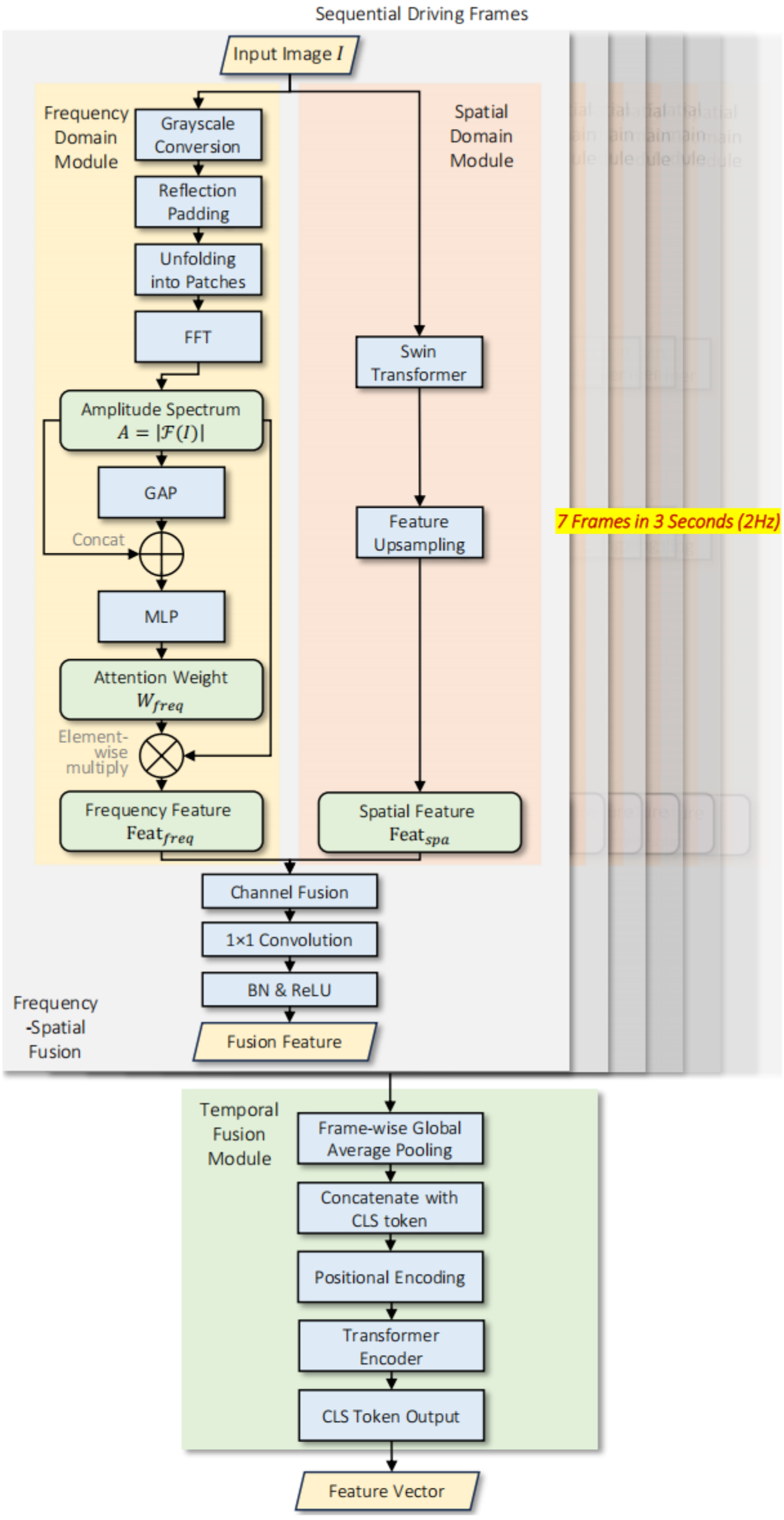

**Fig. 2** The architecture of our proposed TFSF encoder for consecutive driving frames.

For an input RGB image $I_{RGB} \in \mathbb{R}^{3\times H\times W}$, we first convert it into a grayscale representation $I_g \in \mathbb{R}^{1\times H\times W}$ by standard luminance-based weighting:

$$I_g = 0.299 I_R + 0.587 I_G + 0.114 I_B. \tag{1}$$

We partition the grayscale image into non-overlapping patches of size $P \times P$, generating a tensor $X_{patch} \in \mathbb{R}^{N_p\times P\times P}$, where $N_p = \frac{H}{P} \times \frac{W}{P}$. Each patch $X^{(i)}_{patch}$ is independently transformed into the frequency domain via a 2D FFT:

$$F^{(i)} = \mathscr{F}\{X^{(i)}_{\text{patch}}\}, \quad F^{(i)} \in \mathbb{C}^{P\times P}. \tag{2}$$

We subsequently compute the amplitude spectrum $A^{(i)}$:

$$A^{(i)} = |F^{(i)}|, \quad A^{(i)} \in \mathbb{R}^{P\times P}. \tag{3}$$

Then, a global average pooling (GAP) is applied to $A^{(i)}$:

$$a^{(i)} = \text{GAP}(A^{(i)}) = \frac{1}{P^2} \sum_{u=1}^{P} \sum_{v=1}^{P} A^{(i)}_{u,v}, \quad a^{(i)} \in \mathbb{R}. \tag{4}$$

These amplitude statistics across all patches form a vector $a \in \mathbb{R}^{N_p}$. A multilayer perceptron (MLP), parameterized by weights $W_f$ and biases $b_f$, predicts frequency-domain attention weights $w_f \in \mathbb{R}^{N_p}$:

$$w_f = \sigma(\text{MLP}(a; W_f, b_f)), \quad w_f \in [0,1]^{N_p}, \tag{5}$$

where $\sigma$ denotes the sigmoid activation function. The patches are then re-weighted in the spatial domain by multiplying each original grayscale patch $X^{(i)}_{patch}$ with the corresponding scalar attention weight $w^{(i)}_f$:

$$X^{(i)}_f = w^{(i)}_f \cdot X^{(i)}_{\text{patch}}, \quad X^{(i)}_f \in \mathbb{R}^{P\times P}. \tag{6}$$

Finally, we reconstruct the frequency-attended image $X_f \in \mathbb{R}^{1\times H\times W}$ by folding patches back to their original spatial positions.

To effectively capture hierarchical spatial representations, we employ a pre-trained hierarchical Swin Transformer Liu et al. (2021b) (Swin-Tiny) backbone. Given the original RGB image $I_{RGB}$, we feed it into the Swin Transformer:

$$X^{(\ell)}_{\text{spa}} = \text{SwinTransformer}_\ell(I_{RGB}), \quad \ell = 1,2,3,4, \tag{7}$$

where each stage $\ell$ yields feature maps at progressively coarser scales, thus providing multi-scale spatial representations. To align these multi-scale features with the original spatial resolution, we apply bilinear interpolation-based upsampling, resulting in the refined spatial representation $S \in \mathbb{R}^{C\times H\times W}$:

$$X_s = \text{Upsample}\left(\text{Concat}\left(X_{\text{spa}}^{(1)}, X_{\text{spa}}^{(2)}, X_{\text{spa}}^{(3)}, X_{\text{spa}}^{(4)}\right)\right). \tag{8}$$

To integrate complementary information from frequency and spatial domains, we concatenate the frequency-attended image $X_f$ with the spatial features $X_s$ along the channel dimension, yielding the fused tensor $X_{fs} \in \mathbb{R}^{(C+1)\times H\times W}$:

$$X_{fs} = [X_f; X_s]. \tag{9}$$

Subsequently, a $1\times 1$ convolution followed by batch normalization (BN) and ReLU activation generates the unified spatial-frequency representation $X_{feat}$:

$$X_{\text{feat}} = \text{ReLU}\big(\text{BN}\big(\text{Conv}_{1\times 1}(X_{\text{fs}})\big)\big), \quad X_{\text{feat}} \in \mathbb{R}^{C'\times H\times W}. \tag{10}$$

Given a sequence of $T$ consecutive image frames, we encode temporal dependencies among frame-level representations. First, we perform GAP on each spatial-frequency feature map $X_{feat,t}$ (frame $t$), obtaining temporal feature vectors $\mathbf{h}_t \in \mathbb{R}^{C'}$:

$$\mathbf{h}_t = \text{GAP}(X_{\text{feat},t}), \quad t = 1, 2, \ldots, T. \tag{11}$$

The resulting feature vectors across all frames form a temporal feature matrix $\mathbf{H}_{\text{seq}} \in \mathbb{R}^{T\times C'}$:

$$\mathbf{H}_{\text{seq}} = [\mathbf{h}_1; \mathbf{h}_2; \ldots; \mathbf{h}_T]. \tag{12}$$

We prepend a learnable classification token $\mathbf{H}_{\text{cls}} \in \mathbb{R}^{1\times C'}$ and add positional embeddings $\mathbf{P}_{\text{pos}} \in \mathbb{R}^{(T+1)\times C'}$:

$$\mathbf{H}_{\text{in}} = [\mathbf{h}_{\text{cls}}; \mathbf{h}_1; \mathbf{h}_2; \ldots; \mathbf{h}_T] + \mathbf{P}_{\text{pos}}. \tag{13}$$

This sequence is then processed through an $L$-layer Transformer encoder (multi-head self-attention (MHSA) and MLP blocks):

$$\mathbf{H}_{\text{out}} = \text{TransformerEncoder}^{(L)}(\mathbf{H}_{\text{in}}), \quad \mathbf{H}_{\text{out}} \in \mathbb{R}^{(T+1)\times C'}. \tag{14}$$

We take the output corresponding to the classification token as the final feature representation $f_{TFSF}$:

$$\mathbf{h}_{TFSF} = \mathbf{H}_{out}^{(0)}, \quad \mathbf{h}_{TFSF} \in \mathbb{R}^{C'}. \tag{15}$$

For both training and evaluation in this work, we form each input clip by a sliding 3-second window sampled at 2 Hz ($T$ = 7 frames). This makes the sequential input strictly causal and temporally aligned with the reported horizons. Short-horizon planning depends on high-frequency changes that signal relative motion and multi-scale spatial structure that stabilizes geometry over a brief, causal clip. The TFSF encoder explicitly combines FFT-based frequency attention with multi-scale Swin features and a temporal transformer over the 2 Hz window, so we expect these improvements to grow with horizon length (2-3 s) where single-frame cues are relatively weaker.

### 3.2 Self-Supervised Training of the TFSF Encoder

The TFSF encoder is trained through a label-free contrastive learning paradigm that combines in-batch negatives, a fixed-capacity memory queue and a targeted hard-negative mining strategy, as shown in Algorithm 1.

Let $\mathscr{V} = \{V_i\}_{i=1}^{N}$ denote an unlabeled video corpus where each clip $V_i = \{I_{i,t}\}_{t=1}^{T}$ comprises $T$ RGB frames $I_{i,t} \in \mathbb{R}^{3\times H\times W}$. Two independently drawn stochastic augmentations $\mathscr{A}_1$ and $\mathscr{A}_2$ (random-resized crop, horizontal flip, color jitter, Gaussian blur) yield a positive pair:

$$\tilde{V}_i^{(1)} = \mathscr{A}_1(V_i), \quad \tilde{V}_i^{(2)} = \mathscr{A}_2(V_i). \tag{16}$$

The TFSF encoder $f_\theta : \mathbb{R}^{T\times 3\times H\times W} \to \mathbb{R}^d$ maps a clip to a $d$-dimensional representation $\mathbf{h} = f_\theta(\tilde{V})$. A three-layer projection head $\mathscr{G}_\phi : \mathbb{R}^d \to \mathbb{R}^m$ projects $\mathbf{h}$ onto the unit hypersphere:

$$\mathbf{z} = \frac{\mathscr{G}_\phi(\mathbf{h})}{\|\mathscr{G}_\phi(\mathbf{h})\|_2}, \quad \mathbf{z} \in \mathbb{R}^m, \quad d = 64, m = 128. \tag{17}$$

For mini-batch index $i$, we denote the anchor as $\mathbf{z}_i^{(1)}$ and its augmented counterpart as $\mathbf{z}_i^{(2)}$. Their similarity (positive logits) is calculated as follows:

$$s_i^+ = \frac{\mathbf{z}_i^{(1)\top}\mathbf{z}_i^{(2)}}{\tau_{temp}}, \quad \tau_{temp} > 0, \tag{18}$$

**Algorithm 1** Self-supervised training of the TFSF encoder

**Inputs:**

$\mathscr{V}$ — unlabeled video corpus

$\mathscr{E}_\theta, \mathscr{G}_\phi$ — TFSF backbone and projection head

$E, B$ — #epochs, mini-batch size

$\tau_{temp}$ — temperature

$K$ — #hard negatives

$M_Q$ — queue capacity

**Outputs:**

$\theta^*, \phi^*$ — trained parameters

**procedure** TRAIN($\mathscr{V}, \mathscr{E}_\theta, \mathscr{G}_\phi, E, B, \tau_{temp}, K, M_Q$)
    **initialise** $\theta, \phi$ ▷ AdamW defaults
    $Q \leftarrow \emptyset$ ▷ FIFO memory queue
    **for** epoch $\leftarrow 1$ **to** $E$ **do**
        **for** mini-batch $\mathbb{B} \leftarrow$ SAMPLE($\mathscr{V}, B$) **do**
            **for each** clip $V_i \in \mathbb{B}$ **do**
                $V_i^1 \leftarrow \mathscr{A}_1(V_i)$; $V_i^2 \leftarrow \mathscr{A}_2(V_i)$
                $z_i^1 \leftarrow \ell_2_\mathtt{norm}(\mathscr{G}_\phi(\mathscr{E}_\theta(V_i^1)))$
                $z_i^2 \leftarrow \ell_2_\mathtt{norm}(\mathscr{G}_\phi(\mathscr{E}_\theta(V_i^2)))$
            **end for**
            $N \leftarrow \{z_j^k \mid j \neq i \text{ or } k = 2\}_{\forall i} \cup Q$
            **for** $i \leftarrow 1$ **to** $B$ **do**
                $H_i \leftarrow \text{TOP-K}_{\mathtt{roll}}(\text{sim}(z_i^1, \cdot), N, K)$ ▷ K most similar
                $s_i^+ \leftarrow \text{sim}(z_i^1, z_i^2)/\tau_{temp}$
                $\Delta_i \leftarrow \sum_{z \in H_i} \exp(\text{sim}(z_i^1, z)/\tau_{temp})$
                $\mathscr{L}_i \leftarrow -\log\left(\frac{\exp(s_i^+)}{\exp(s_i^+) + \Delta_i}\right)$
            **end for**
            $\mathscr{L} \leftarrow (1/B)\sum_i \mathscr{L}_i$
            $(\theta, \phi) \leftarrow$ ADAMW($(\theta, \phi), \nabla \mathscr{L}$)
            $Q \leftarrow$ ENQUEUE($Q, \{z_i^2\}$); TRIM($Q, M_Q$)
        **end for**
    **end for**
    **return** $\theta^*, \phi^*$
**end procedure**

**Notation:** $\text{sim}(u, v) = u^\top v$; $\ell_2_\mathtt{norm}(x) = x/\|x\|_2$; $\text{TOP-K}_{\mathtt{roll}}$ returns the $K$ largest inner-products.

where two negative pools are employed, namely the in-batch negatives $\mathcal{N}_i^{\mathrm{B}} = \left\{ \mathbf{z}_j^{(k)} \mid j \neq i \text{ or } k = 2 \right\}$ and the memory-queue negatives $\mathcal{Q} = \{\mathbf{q}_\ell\}_{\ell=1}^{M}$ with fixed size $M = 1024$, storing projections from recent mini-batches.

Since most negatives are easily separable, we retain only the $K$ most similar hard negatives:

$$\mathcal{N}_i^* = \mathrm{TopK}\left(\mathcal{N}_i^{\mathrm{B}} \cup \mathcal{Q}, \mathbf{z}_i^{(1)\top}\mathbf{z}\right), \quad K = 10. \tag{19}$$

The InfoNCE loss could be computed as the following formula:

$$\mathcal{L}_i = -\log \frac{\exp(s_i^+)}{\exp(s_i^+) + \sum_{\mathbf{z} \in \mathcal{N}_i^*} \exp\left(\frac{\mathbf{z}_i^{(1)\top}\mathbf{z}}{\tau_{temp}}\right)}, \tag{20}$$

and the batch loss could be defined as $\mathcal{L} = B^{-1} \sum_{i=1}^{B} \mathcal{L}_i$.

After each iteration the vectors $\left\{ \mathbf{z}_i^{(2)} \right\}_{i=1}^{B}$ are en-queued and the oldest $B$ items are dequeued, keeping $|\mathcal{Q}| = M$. Both the backbone parameters $\theta$ and the projection parameters $\phi$ receive direct gradient updates:

$$\theta \leftarrow \theta - \eta_e \nabla_\theta \mathcal{L}, \quad \phi \leftarrow \phi - \eta_p \nabla_\phi \mathcal{L}, \tag{21}$$

with learning rates $\eta_e = 1 \times 10^{-5}, \quad \eta_p = 1 \times 10^{-4}$ and weight decay $1 \times 10^{-4}$. Training is conducted for $E = 50$ epochs on 7-frame clips, batch size $B = 8$ and temperature $\tau_{temp} = 0.07$ on a single 3090 GPU. The model with minimum validation loss is preserved for downstream embedding and retrieval tasks.

### 3.3 The Embedding and Retrieval Pipeline

The overview of the embedding and retrieval pipeline is shown in Algorithm 2. Given a collection of unlabeled video sequences, a database embedding set could be constructed firstly using the trained TFSF encoder coupled with a projection head. Specifically, for each video clip $V_i = \{I_{i,t}\}_{t=1}^{T}$, containing $T = 7$ consecutive frames, we perform deterministic image preprocessing, including resizing to $224 \times 224$, normalization by ImageNet mean and standard deviation, and tensor stacking. These preprocessed frames form an input tensor $X_i \in \mathbb{R}^{1 \times 7 \times 3 \times 224 \times 224}$.

Each video sequence embedding is computed as:

$$\mathbf{h}_i = \mathcal{E}_\theta(X_i), \quad \mathbf{e}_i = \frac{\mathcal{G}_\phi(\mathbf{h}_i)}{\|\mathcal{G}_\phi(\mathbf{h}_i)\|_2} \in \mathbb{R}^{128}, \tag{22}$$

where $\mathcal{E}_\theta$ denotes the trained TFSF encoder and $\mathcal{G}_\phi$ represents a multilayer projection head consisting of linear layers, batch normalization, and ReLU activations. This produces an embedding $\mathbf{e}_i$ that resides on the 128-dimensional unit hypersphere. Embeddings for all sequences are subsequently stored in a database along with unique identifiers for retrieval tasks.

To facilitate efficient retrieval, we adopt a two-stage indexing strategy combining unsupervised clustering with hierarchical navigable small-world graphs (HNSW) Malkov and Yashunin (2018).

Given the database embeddings $\mathcal{E}_{\text{db}} = \{\mathbf{e}_i\}_{i=1}^{N_{\text{db}}}$, we first partition them into $n_c$ clusters via k-means clustering:

$$\arg\min_{\{\mathbf{c}_j\}} \sum_{i=1}^{N_{\text{db}}} \min_j \|\mathbf{e}_i - \mathbf{c}_j\|_2^2, \tag{23}$$

where $\{\mathbf{c}_j\}_{j=1}^{n_c}$ denote cluster centroids. Each embedding is thus associated with a cluster label.

Within each cluster, we construct a separate HNSW index designed for efficient approximate nearest-neighbor search under cosine distance:

$$d_{\cos}(\mathbf{x}, \mathbf{y}) = 1 - \frac{\mathbf{x}^\top \mathbf{y}}{\|\mathbf{x}\|_2 \|\mathbf{y}\|_2}. \tag{24}$$

Specifically, for cluster $j$, an HNSW index is initialized and built using all embeddings within that cluster. Parameters including $M = 16$ and ef_construction $= 200$ are set to balance retrieval speed and accuracy.

**Algorithm 2** Embedding and Retrieval Pipeline

**Inputs:**

$V_{\text{db}}, V_{\text{val}}$ — Database and validation video clips

$\mathscr{E}_\theta, \mathscr{G}_\phi$ — Trained TFSF encoder and projection head

$\text{nc}, M$ — Number of clusters, HNSW connectivity

**procedure** EMBEDDING AND RETRIEVAL
    **1. Embedding Generation**
    **for** each clip $V_i \in V_{\text{db}}$ **do**
        $X_i \leftarrow \text{Preprocess}(V_i)$ ▷ deterministic transformations
        $h_i \leftarrow \mathscr{E}_\theta(X_i)$
        $e_i \leftarrow \text{Normalize}(\mathscr{G}_\phi(h_i))$
    **end for**
    Save embeddings $\{e_i\} \rightarrow$ database
    **2. Clustered HNSW Indexing**
    $\{c_j\}, \text{labels} \leftarrow \text{k-means}(\{e_i\}, \text{n_clusters=nc})$
    **for** each cluster $j$ **do**
        Build $\text{HNSW}(\{e_i \mid \text{labels}[i] = j\}, \text{connectivity} = M)$
    **end for**
    **3. Retrieval**
    **for** each validation clip $V_k \in V_{\text{val}}$ **do**
        $X'_k \leftarrow \text{Preprocess}(V_k)$
        $h'_k \leftarrow \mathscr{E}_\theta(X'_k), e'_k \leftarrow \text{Normalize}(\mathscr{G}_\phi(h'_k))$
        $c_j \leftarrow \text{k-means.predict}(e'_k)$
        $\text{retrieved_indices} \leftarrow \text{HNSW_query}(e'_k, \text{cluster} = c_j, \text{topk} = k)$
        retrieved_trajectories $\leftarrow$ Map indices $\rightarrow$ trajectories
        Output retrieved_trajectories
    **end for**
**end procedure**

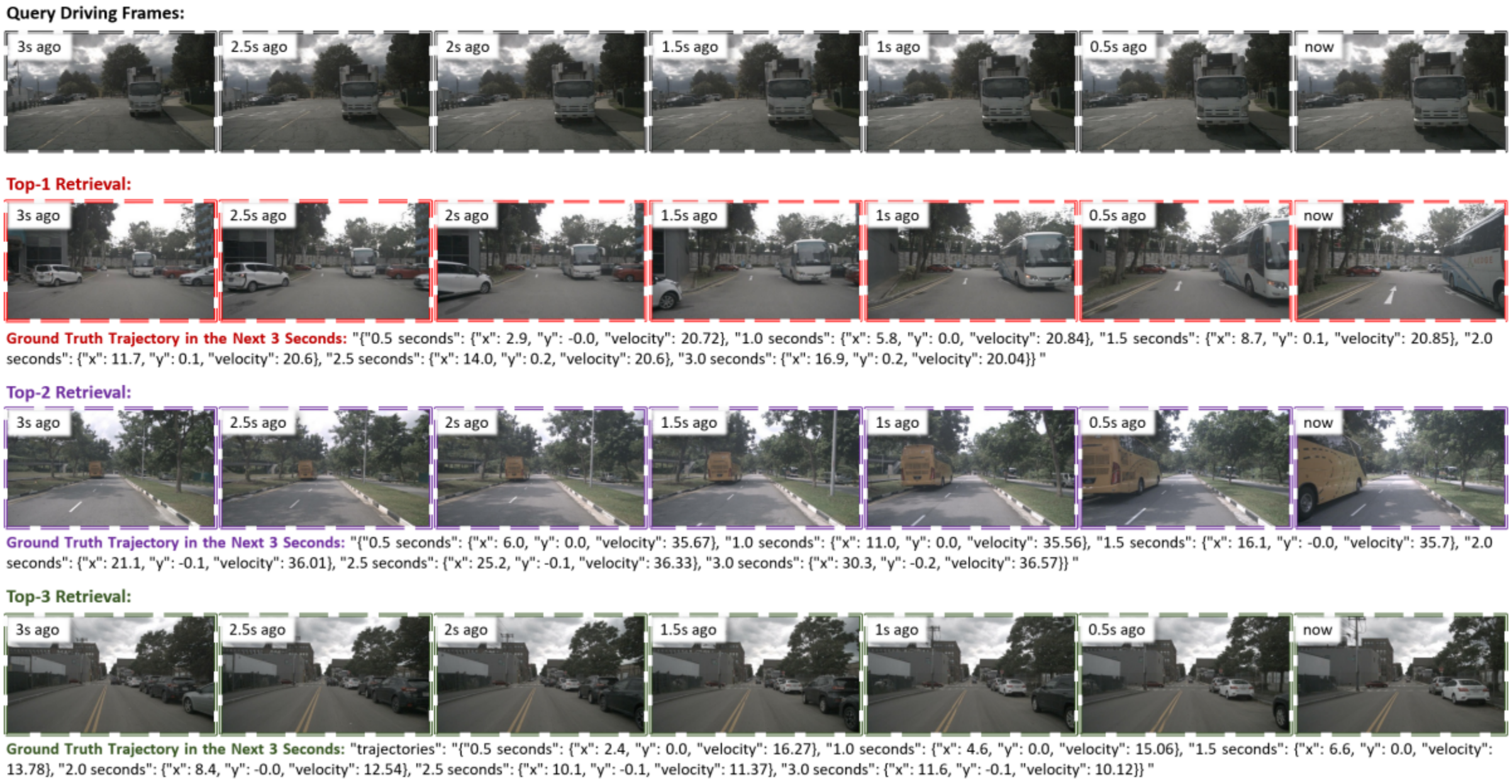

**Fig. 3** Illustration of the embedding-based retrieval pipeline in action.

Given validation embeddings $\mathscr{E}_{\text{val}} = \left\{\mathbf{e}'_k\right\}_{k=1}^{N_{\text{val}}}$, retrieval is conducted by predicting cluster assignments via the pre-fitted k-means model, and querying the corresponding cluster-specific HNSW index to identify the $k$ nearest database embeddings:

$$\{\mathbf{e}_l\}_{l=1}^{k} = \text{HNSW_knn_query}(\mathbf{e}'_k). \tag{25}$$

Each retrieved embedding index corresponds to an entry in the original database JSON file, from which associated ground truth trajectories are extracted. Formally, given validation embedding index $k$ and retrieved indices $\{l_1, \ldots, l_k\}$, the output structure is:

$$R_k = \left\{\mathscr{T}(l_j)\right\}_{j=1}^{k}, \tag{26}$$

where $\mathscr{T}(l_j)$ denotes the trajectory annotation corresponding to database embedding index $l_j$. This is a pipeline designed for large-scale validation datasets. In practical applications, a new scene sequence can be independently embedded and used for one-to-one retrieval, retaining a single trajectory.

Representative retrieval examples are illustrated in Figure 3. The top row displays a sequence of query driving frames sampled at uniform intervals over the past 3 seconds, capturing a scenario involving a truck ahead. The subsequent three rows illustrate the top-3 retrieval results obtained by the proposed embedding and retrieval method. Each retrieval result consists of a visually similar sequence of frames, retrieved from the training database using the learned embeddings. The ground truth trajectories corresponding to each retrieved sequence are provided, including position coordinates (x, y) and velocities (in km/h) for future timestamps at intervals of 0.5 seconds, spanning the next 3 seconds. These retrieved cases demonstrate the ability of the embedding pipeline to effectively capture visual similarity and retrieve semantically relevant driving scenarios, thereby providing useful trajectory priors for downstream prediction tasks.

## 3.4 The Triple-Stage Fine-Tuning Strategy

Reliable trajectory prediction in autonomous driving relies on VLMs' ability to accurately perceive spatial relationships in complex traffic scenes. Although VLMs have already demonstrated strong reasoning capabilities, their depth and scale estimation from monocular inputs often remains suboptimal, particularly for distant targets. Such deficiencies can cascade through the prediction pipeline, leading to unsafe trajectories. Therefore, we adopt a triple-stage fine-tuning strategy for VLMs, in which the first stage is designed to enhance spatial perception capabilities from front-view frames, while the second and the third stages optimize trajectory prediction using surround-view images and consecutive front-view frames respectively, along with ground-truth trajectories.

### 3.4.1 Stage A: Towards Enhancement of Spatial Perception Capabilities of VLMs

In this stage, the vision encoder is frozen, and only the LLM backbone of the VLM is adapted to enhance spatial perception. We construct a spatial perception dataset from nuScenes in a VQA format, containing over 100,000 verified question–answer pairs that cover three perception tasks essential for planning: object classification, physical size estimation, and distance inference Wang et al. (2025c). Examples of distance inference tasks in the training dataset are illustrated in Figure 4.

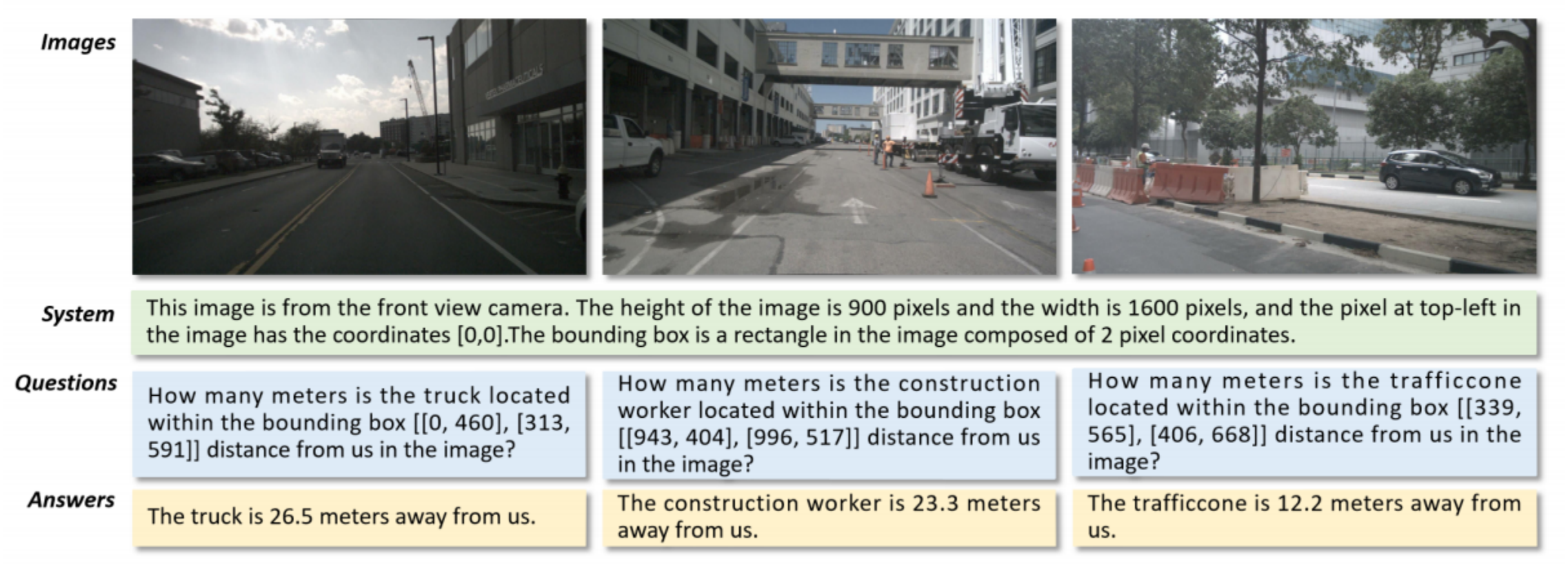

**Fig. 4** Examples of distance inference tasks in the training dataset.

The multi-task loss for this stage is formulated as:

$$\mathscr{L}_{\mathrm{A}} = \frac{1}{B}\sum_{t=1}^{B}\left[\alpha_t \cdot \left(-\sum_{k=1}^{K}\eta_{k,t}\log\pi_{k,t}\right) + \beta_t \cdot \frac{1}{3}\sum_{m=1}^{3}(s_{m,t}-\hat{s}_{m,t})^2 + \gamma_t \cdot (d_t - \hat{d}_t)^2\right], \tag{27}$$

where $B$ denotes the batch size, $K$ is the number of semantic categories, $\eta_{k,t}$ and $\pi_{k,t}$ represent the one-hot ground-truth label and predicted probability for category $k$ in sample $t$, $s_{m,t}$ and $\hat{s}_{m,t}$ indicate the predicted and reference object size along dimension $m$, $d_t$ and $\hat{d}_t$ refer to the predicted and actual distance to the ego vehicle, and $\alpha_t$, $\beta_t$, $\gamma_t$ are binary indicators specifying whether the classification, size regression, and distance regression terms are active for that sample.

We adopt Qwen-series VLMs as the backbone, and use LoRA Hu et al. (2022a) as the fine-tuning framework. This stage aims to enhance the VLMs' spatial perception ability, therefor forming a robust perceptual foundation for the subsequent fine-tuning.

#### 3.4.2 Stage B: Towards Scene-Conditioned Trajectory Prediction from Surround-View Images

Stage B is designed to train VLMs to output trajectories from six input surround-view images. Each training instance includes the surround-view images, along with the ego vehicle's current status, the historical trajectory over the past three seconds, and a guiding way-point that indicates the intended direction within the prediction horizon.

Visual tokens are extracted by a frozen vision encoder and fused with the current velocity, the past three-second trajectory, and the way-point before entering the LLM backbone. The training objective fits coordinates and velocities through a standard regression loss, with simple regularizers that encourage smooth motion and reasonable curvature over time:

$$\mathscr{L}_{\mathrm{B}} = \frac{1}{B}\sum_{r=1}^{B}\frac{1}{Q}\sum_{q=1}^{Q}\left[(x_{q,r}-\tilde{x}_{q,r})^2+(y_{q,r}-\tilde{y}_{q,r})^2+(v_{q,r}-\tilde{v}-_{q,r})^2\right], \tag{28}$$

where $B$ denotes the batch size, $Q = 3$ denotes the prediction horizons (1s, 2s, 3s), $(x_{q,r}, y_{q,r}, v_{q,r})$ are the predicted ego-centric x/y coordinates and velocity for sample $r$ at horizon $q$, tildes indicate their ground-truth counterparts. The LoRA fine-tuning is still applied in this stage.

This stage trains VLMs to generate physically realizable, short-term trajectories according to surround-view images. The resulting motion-aligned representation forms the bridge from Stage A's perception enhancement to Stage C's end-to-end trajectory prediction.

#### 3.4.3 Stage C: Towards End-to-End Trajectory Prediction from Consecutive Front-View Frames

Stage C finalizes the triple-stage pipeline by training VLMs to produce short-horizon ego trajectories directly from a consecutive front-view frame sequence. Each training instance supplies seven front-view frames timestamped at $-3.0, -2.5, -2.0, -1.5, -1.0, -0.5, 0.0$ seconds relative to the present, a short history of ego positions and velocities expressed in the ego frame, and an optional soft way-point that provides a directional prior without imposing arrival within the prediction horizon.

Architecturally, the seven frames are tokenized by the frozen vision encoder and augmented with relative-time embeddings to preserve temporal order. These visual tokens are concatenated with structured kinematic features (history trajectory way-points and velocities). Consistent with earlier stages, we retain a Qwen-series VLM and adapt only the LLM parameters via LoRA, which keeps adaptation parameter-efficient while focusing the optimization on temporal reasoning from consecutive frames.

To avoid mathematical duplication, Stage C reuses the trajectory regression objective introduced in Stage B ($\mathscr{L}_{\mathrm{C}} = \mathscr{L}_{\mathrm{B}}$), which jointly penalizes coordinate and velocity errors. In practice, Stage C minimizes $\mathscr{L}_{\mathrm{C}}$ without introducing new loss terms, only task-specific tuning of the regularization weights is performed to accommodate the reduced field-of-view and the stronger reliance on temporal cues.

Stage C therefore completes this triple-stage fine-tuning strategy by converting the perception ca-pabilities from Stage A and the surround-view priors from Stage B into a front-view–centric temporal planner. Stage C integrates seamlessly into the unified optimization framework while contributing front-view–specific temporal reasoning essential for practical, deployable single-camera planning.

#### 3.4.4 Practical Choice between LoRA and GRPO

While reinforcement learning (RL)-based preference optimization methods, such as group relative policy optimization (GRPO), have recently shown promise for aligning VLMs, our design objective in this work is to optimize short-horizon trajectory coordinates with physics-aware regularization and safety proxies under a RAG framework. In this setting, LoRA-based fine-tuning of the LLM backbone of a VLM offers three practical advantages:

**Objective alignment and stability.** Our loss directly matches the evaluation metrics (L2 displacement and collision rate). Supervised, gradient-based training with LoRA offers a straightforward and stable optimization process. In contrast, using GRPO-style reinforcement learning would require to design surrogate rewards for trajectories, to handle temporal credit assignment over long horizons, and to carefully tune safety-related factors, which are all known to introduce instability and high variance in practice.

**Data efficiency and modularity.** LoRA updates only a small set of parameters and fits neatly into our triple-stage fine-tuning pipeline, without needing to unfreeze the vision encoders or retrain any reward models. This makes the training process parameter-efficient, easy to reproduce, and well aligned with the deployment-latency requirements of our retrieval pipeline.

For these reasons, we use LoRA in this work because it aligns well with our metric-supervised objectives and deployment needs, and it provides a stable and low-variance way to optimize the model. GRPO-style alignment is still valuable, but mainly for future efforts that focus on preference- or rule-based refinements such as comfort, driving style, or normative compliance, which are outside the scope of our current open-loop evaluation.

### 3.5 The Chain-of-Thought Prompting Paradigm

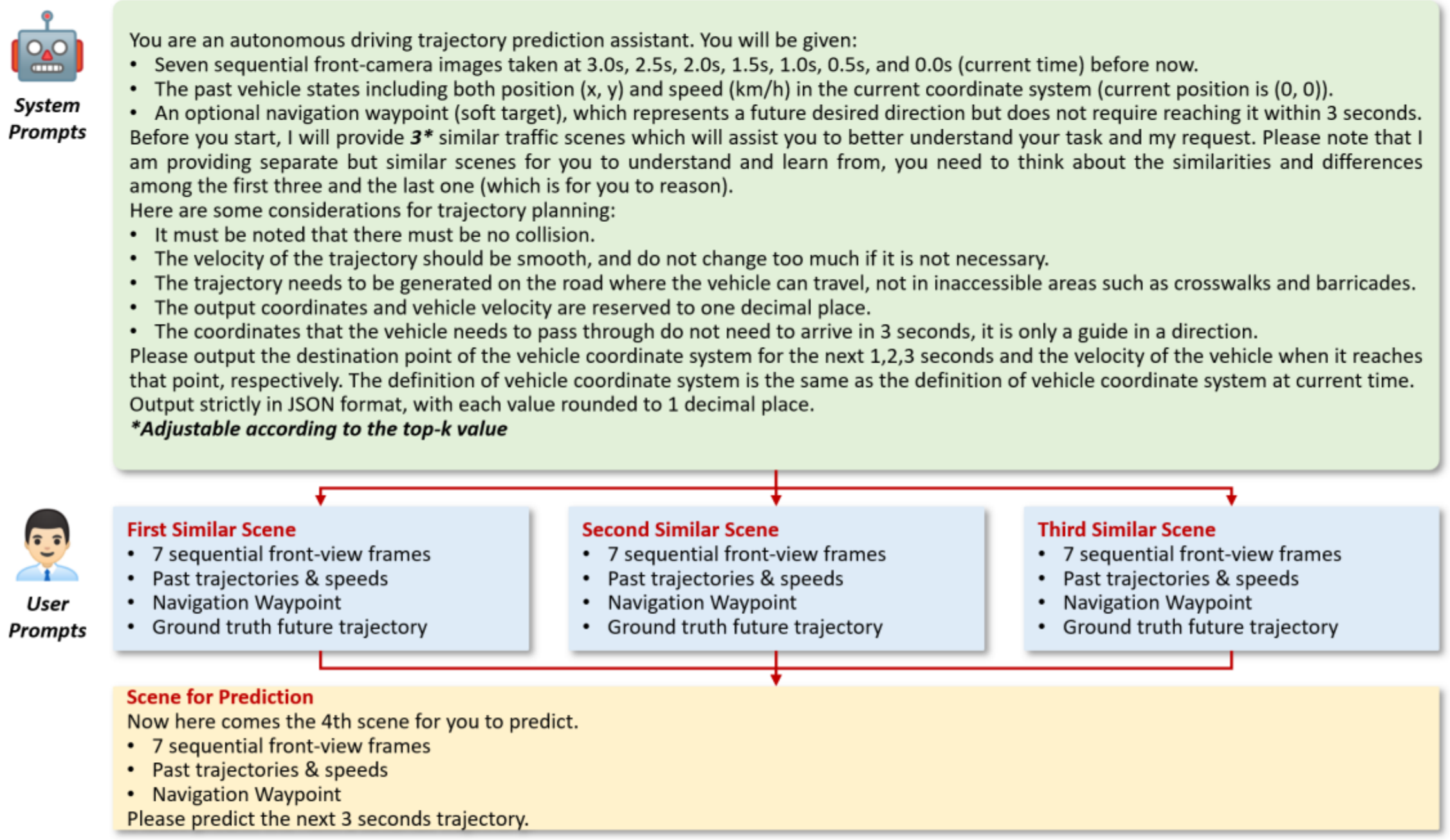

**Fig. 5** An example of the CoT prompting paradigm.

To strengthen the reasoning capabilities of VLMs in trajectory prediction for autonomous driving, we employ a CoT prompting strategy that encourages the VLM to reason step by step using retrieved reference scenes. Instead of relying on standard prompts that ask for a prediction based solely on the current scene, our method incorporates structured multi-modal context and comparative reasoning so that the VLM can follow a more human-like decision process.

As illustrated in Figure 5, our CoT prompting is organized around a multi-scene structure. The model is first presented with several reference scenes that are retrieved through our embedding and retrieval pipeline, where the number of retrieved samples is controlled by the Top-K setting. After processing these reference scenes, the model receives the target scene for inference. Each reference scene contains the following components:

- A sequence of seven front-view images captured at time steps {3.0, 2.5, 2.0, 1.5, 1.0, 0.5, 0.0} seconds before the present,
- historical trajectory data including position and velocity in the ego vehicle coordinate system,

- an optional navigation way-point representing a desired future direction,
- the ground truth future trajectory of the ego vehicle.

This exemplar-guided context helps the model form a clear understanding of how scene dynamics relate to vehicle motion and the types of future trajectories that are feasible. By examining multiple instances, the model learns to recognize recurring patterns, interpret scene-specific constraints, and transfer this understanding to new but related driving scenarios. To ensure coherent and safe trajectory generation, the system prompt specifies several concrete planning considerations, such as collision avoidance, velocity smoothness, and adherence to drivable areas. These considerations improve the feasibility of outputs.

A key feature of our CoT paradigm is its focus on relational reasoning. The model is guided to compare the reference scenes with the target scene and to identify both their common structure and distinguishing characteristics. This strengthens the transfer of knowledge across scenes and supports effective contextual adaptation. In addition, all spatial information is represented in a unified vehicle-centric coordinate system, and temporal cues are explicitly annotated, which provides clear temporal and spatial grounding for the trajectory prediction task.

Similarly, the prompt structure is designed so that the ego vehicle's dynamic status, such as velocity, acceleration, and yaw angle, could be incorporated directly during inference. These values could be supplied to the system without extra pre-processing, which allows the model to make more accurate and context-wise decisions.

# 4 Experiment

## 4.1 Experimental Setup

### 4.1.1 Data Preparation

We divide the 34,000 scenes from the nuScenes dataset Caesar et al. (2020) into four subsets:

- The first subset consists of 10,000 scenes. After removing redundant frames that lack either the

first 3 seconds of historical context or the last 3 seconds of future trajectory information, a total of 9,062 overlapping scene sequences remain. We use this subset to train the TFSF vision encoder.

- The second subset also contains 10,000 scenes. In Stage A, we use the surround-view images from this subset to build a fine-tuning dataset to enhance VLMs' spatial perception abilities. In Stage B, we further fine-tune VLMs using both the surround-view images and corresponding 3-second future trajectory ground truths, enhancing its trajectory prediction performance. Following the same filtering strategy as in the first subset, front-view frames of these 10,000 scenes are divided into 1,117 non-overlapping scene sequences. These sequences, paired with their 3-second future trajectory ground truths, are utilized in Stage C, serving as the final stage to jointly optimize spatiotemporal reasoning and trajectory prediction.
- The third subset also includes 10,000 scenes. As with the first subset, we extract 9,062 overlapping scene sequences. Each sequence is paired with its corresponding 3-second future trajectory and encoded into a joint vector representation with the TFSF encoder. These representations are stored in a vector-based knowledge base, which is used in the RAG pipeline.
- The remaining 4,000 scenes are used to construct the validation set, from which 3,871 valid overlapping scene sequences are obtained. This subset is used to evaluate the overall performance of VLMs.

All training or fine-tuning subsets (for the TFSF encoder training, Stage-A/B/C fine-tuning, and building the database) are constructed exclusively from the nuScenes training data and are disjoint from the held-out official test set. All comparative results reported in Section 4.2 are obtained on the nuScenes official test split.

#### 4.1.2 Evaluation Metrics

The prediction of trajectories is evaluated using the L2 displacement error and the collision rate. More specifically, we adopt the L2 and collision rate based on UniAD Hu et al. (2023) and ST-P3 Hu et al. (2022b). UniAD's **NoAvg** protocol computes the error at each horizon using only the prediction made for that specific horizon, without any temporal averaging. In contrast, ST-P3's **TemAvg** protocol computes the score at a horizon as the cumulative arithmetic mean of the intermediate scores of 0.5s up to that horizon, which smooths the results over time. In the comparative experiments, we report KEPT's performance under both protocols so that it is directly comparable with different baselines. In the ablation study, since external comparisons are not required, we evaluate performance using the NoAvg protocol.

#### 4.1.3 The Vision-Language Model Backbone

In our experiments, we adopt the Qwen2-VL-2B model as the default VLM backbone. As previously noted, this component is modular and can be substituted with other models in the Qwen family, including Qwen2-VL-7B, Qwen2.5-VL-3B, Qwen2.5-VL-7B, or larger-scale variants. We choose the 2B model because it requires far less computation during both training and inference. This makes it suitable for real-time use and for deployment on resource-limited platforms. A further ablation study on various VLM backbones is conducted in Section 4.3.5. All evaluation is conducted on four NVIDIA A800 GPUs.

### 4.2 Comparative Experiments

All comparisons are performed on the official nuScenes test split using a shared open-loop setting. Baselines that operate on single-frame vision inputs are evaluated directly on the same test scenes without retraining. In contrast, KEPT processes a rolling three-second clip sampled at 2 Hz, which provides seven frames, and we report both the NoAvg (per-horizon) and TemAvg (temporal average up to the horizon) results to match common reporting conventions. Methods that make use of ego status information such as velocity, acceleration, or yaw angle are marked with an asterisk in Table 2. We provide the results of KEPT under both configurations to maintain fair comparisons. This evaluation setup preserves the intended usage of each baseline and highlights the specific benefit introduced by KEPT's short-horizon temporal context.

The methods summarized in Table 2 fall into four representative families, each incorporating geometry, temporal cues and semantics into the planning process in a different way. End-to-end BEV-centric planners such as UniAD Hu et al. (2023), ST-P3 Hu et al. (2022b), and BEV-Planner Li et al. (2024) transform multi-view images into BEV features and decode way-points within a unified planning network.

| NoAvg | TemAvg | Methods | L2 (m)↓ | | | | Collision (%)↓ | | | |
|---|---|---|---|---|---|---|---|---|---|---|
| | | | 1s | 2s | 3s | Avg. | 1s | 2s | 3s | Avg. |
| × | × | FF Hu et al. (2021b) | 0.55 | 1.20 | 2.54 | 1.43 | 0.06 | 0.17 | 1.07 | 0.43 |
| × | × | EO Khurana et al. (2022) | 0.67 | 1.36 | 2.78 | 1.60 | 0.04 | 0.09 | 0.88 | 0.33 |
| ✓ | × | ST-P3 Hu et al. (2022b) | 1.72 | 3.26 | 4.86 | 3.28 | 0.44 | 1.08 | 3.01 | 1.51 |
| ✓ | × | OccNet Ahmed et al. (2023) | 1.29 | 2.13 | 2.99 | 2.14 | 0.21 | 0.59 | 1.37 | 0.72 |
| ✓ | × | UniAD Hu et al. (2023) | 0.48 | 0.96 | 1.65 | 1.03 | **0.05** | 0.17 | 0.71 | 0.31 |
| ✓ | × | VAD-Base Jiang et al. (2023a) | 0.54 | 1.15 | 1.98 | 1.22 | 0.10 | 0.24 | 0.96 | 0.43 |
| ✓ | × | Drive-OccWorld Yang et al. (2025b) | 0.32 | 0.75 | 1.49 | 0.85 | **0.05** | 0.17 | 0.64 | 0.29 |
| ✓ | × | **KEPT (ours, on Qwen2-VL-2B)** | **0.23** | **0.63** | **1.23** | **0.70** | **0.05** | **0.14** | **0.44** | **0.21** |
| × | ✓ | ST-P3 Hu et al. (2022b) | 1.33 | 2.11 | 2.90 | 2.11 | 0.23 | 0.62 | 1.27 | 0.71 |
| × | ✓ | UniAD Hu et al. (2023) | 0.44 | 0.67 | 0.96 | 0.69 | 0.04 | **0.08** | 0.23 | 0.12 |
| × | ✓ | VAD-Base Jiang et al. (2023a) | 0.41 | 0.70 | 1.05 | 0.72 | 0.07 | 0.17 | 0.41 | 0.22 |
| × | ✓ | Drive-OccWorld Yang et al. (2025b) | 0.25 | 0.44 | 0.72 | 0.47 | **0.03** | **0.08** | **0.22** | **0.11** |
| × | ✓ | **KEPT (ours, on Qwen2-VL-2B)** | **0.17** | **0.35** | **0.59** | **0.37** | **0.03** | 0.10 | 0.34 | 0.16 |
| ✓ | × | GPT-Driver* Mao et al. (2023) | 0.27 | 0.74 | 1.52 | 0.84 | 0.07 | 0.15 | 1.10 | 0.34 |
| ✓ | × | RDA-Driver* Huang et al. (2024b) | 0.23 | 0.73 | 1.54 | 0.80 | **0.00** | 0.13 | 0.83 | 0.32 |
| ✓ | × | DME-Driver* Han et al. (2025) | 0.43 | 0.91 | 1.58 | 0.97 | 0.04 | 0.14 | 0.64 | 0.27 |
| ✓ | × | VLP* Pan et al. (2024) | 0.36 | 0.68 | 1.19 | 0.74 | 0.03 | 0.12 | 0.32 | 0.16 |
| ✓ | × | **KEPT* (ours, on Qwen2-VL-2B)** | **0.19** | **0.52** | **1.12** | **0.61** | 0.02 | **0.11** | **0.29** | **0.14** |
| × | ✓ | UniAD* Hu et al. (2023) | 0.20 | 0.42 | 0.75 | 0.46 | 0.02 | 0.25 | 0.84 | 0.37 |
| × | ✓ | VAD-Base* Jiang et al. (2023a) | 0.17 | 0.34 | 0.60 | 0.37 | 0.04 | 0.27 | 0.67 | 0.33 |
| × | ✓ | GPT-Driver* Mao et al. (2023) | 0.20 | 0.40 | 0.70 | 0.44 | 0.04 | 0.12 | 0.36 | 0.17 |
| × | ✓ | RDA-Driver* Huang et al. (2024b) | 0.17 | 0.37 | 0.69 | 0.40 | 0.01 | **0.05** | 0.26 | 0.10 |
| × | ✓ | BEV-Planner* Li et al. (2024) | 0.16 | 0.32 | 0.57 | 0.35 | **0.00** | 0.29 | 0.73 | 0.34 |
| × | ✓ | VLP* Pan et al. (2024) | 0.30 | 0.53 | 0.84 | 0.55 | 0.01 | 0.07 | 0.38 | 0.15 |
| × | ✓ | DriveVLM* Tian et al. (2025) | 0.18 | 0.34 | 0.68 | 0.40 | 0.10 | 0.22 | 0.45 | 0.27 |
| × | ✓ | OmniDrive* Wang et al. (2025a) | **0.14** | 0.29 | 0.55 | 0.33 | **0.00** | 0.13 | 0.78 | 0.30 |
| × | ✓ | Drive-OccWorld* Yang et al. (2025b) | 0.17 | 0.31 | **0.49** | 0.32 | 0.02 | 0.24 | 0.62 | 0.29 |
| × | ✓ | **KEPT* (ours, on Qwen2-VL-2B)** | **0.14** | **0.28** | 0.51 | **0.31** | 0.01 | 0.06 | **0.13** | **0.07** |

**Table 2** Performance comparison with representative published baselines and our KEPT. Results from pre-prints are excluded. ↓ indicates ”lower is better”. $*$ indicates the use of ego status as inputs.

These methods are closeling tied to the planning objective, yet they usually rely on single-frame inputs during inference, which could make long-horizon behavior sensitive to momentary visual noise. Vectorized approaches like VAD-Base Jiang et al. (2023a) model lanes and agents as structured vectors, which serve as geometric constraints, making the planner more disciplined and more efficient. Occupancy and world-model pipelines, such as OccNet Ahmed et al. (2023) and Drive-OccWorld Yang et al. (2025b), take a different route. They reconstruct or predict the multi-dimensional occupancy of the environment before generating a plan. This design provides clear geometric grounding, although it increases computation and extends the planning chain. The last group of methods uses VLMs or LLMs, including GPT-Driver Mao et al. (2023), RDA-Driver Huang et al. (2024b), DME-Driver Han et al. (2025), DriveVLM Tian et al. (2025), and OmniDrive Wang et al. (2025a). These models bring language-based reasoning into the planning process, which improves interpretability and supports higher-level decision making. They often require additional components to connect their text-based reasoning to accurate geometric constraints.

From this perspective, KEPT follows a different design strategy. It introduces short-horizon temporal fusion at the input stage through the TFSF encoder, incorporates compact retrieval priors by RAG, and performs decoding with CoT prompts. With these design choices, the results in Table 2 form a clear trend. KEPT performs better under both the NoAvg and TemAvg protocols. This suggests that its improvements reflect real gains in trajectory quality. Temporal averaging mainly smooths out frame-to-frame noise. Since KEPT already takes in temporally coherent inputs, averaging simply make this stability more visible, especially when compared with single-frame baselines.

At the same time, the largest performance gaps in Table 2 appear at the far-horizon range of 2 to 3 seconds. This is exactly where single-frame BEV planners tend to fail, especially under low visibility or partial occlusion. In such conditions, some water reflections or shadows could be mistaken for real structure and cause lateral drift. Also, the lack of historical motion cues could delay longitudinal responses. KEPT addresses these issues in two complementary ways. The TFSF encoder extracts features from a short rolling clip, which strengthens signals that remain stable across frames, while reducing the influence of transient artifacts. The RAG pipeline then provides additional prior knowledge to help stabilize the decoder when the visual evidence becomes sparse. By the 2 to 3 second horizon, this combination produces trajectories that stay closer to the ground truth and maintain a smoother velocity pattern.

Moreover, comparisons within each method family help clarify practical trade-offs. Compared with occupancy and world-model pipelines, KEPT retains temporal consistency, and also avoids the heavy cost of multi-dimensional forecasting and the long generative chain. This makes KEPT more suitable for deployment under realistic compute limits. Compared with other VLM-based planners, KEPT gives up detailed natural-language explanations in exchange for better spatial grounding. This trade-off leads to higher-quality open-loop way-points.

In conclusion, KEPT's performance should be understood not as a small leaderboard gain but as evidence that combining short-horizon temporal fusion with RAG and CoT leads to more stable far-horizon trajectories across a range of methods on a shared benchmark.

## 4.3 Ablation Studies

### 4.3.1 Ablation Study on Fine-Tuning Stages

| **Fine-Tuning Stages** | | | **Ego Status** | **L2 (m)↓** | | | | **Collision (%)↓** | | | |
|---|---|---|---|---|---|---|---|---|---|---|---|
| **Stage A** | **Stage B** | **Stage C** | **Current** ($v, a, yaw$) | 1s | 2s | 3s | Avg. | 1s | 2s | 3s | Avg. |
| × | ✓ | ✓ | × | 0.23 | 0.65 | 1.40 | 0.76 | 0.15 | 0.39 | 0.89 | 0.48 |
| ✓ | × | ✓ | × | 0.23 | 0.69 | 1.41 | 0.78 | 0.13 | 0.42 | 0.93 | 0.49 |
| ✓ | ✓ | ✓ | × | **0.23** | **0.64** | **1.25** | **0.71** | **0.06** | **0.16** | **0.51** | **0.24** |
| × | ✓ | ✓ | ✓ | 0.23 | 0.64 | 1.39 | 0.75 | 0.14 | 0.37 | 0.83 | 0.45 |
| ✓ | × | ✓ | ✓ | 0.22 | 0.66 | 1.36 | 0.75 | 0.11 | 0.44 | 0.97 | 0.51 |
| ✓ | ✓ | ✓ | ✓ | **0.21** | **0.56** | **1.13** | **0.63** | **0.03** | **0.13** | **0.31** | **0.16** |

**Table 3** Ablation study on fine-tuning stages. Only the NoAvg protocol is adopted. ↓ indicates lower is better. The Top-K value of RAG is set to 1 accordingly. The VLM backbone is Qwen2-VL-2B.

We assess the contribution of the three fine-tuning stages by selectively enabling or disabling Stage A and Stage B while keeping Stage C active. The numerical results are provided in Table 3. As discussed in Section 3.4, Stage A enhances spatial perception, Stage B adds motion reasoning with surround-view images, and Stage C directly optimizes trajectory prediction on consecutive front-view frames.

**(i) Without ego status.** Enabling all three stages produces the best accuracy and safety, with an

average L2 error of 0.71m and an average collision rate of 0.24%. When Stage A is removed, the average L2 increases to 0.76m and the collision rate rises to 0.48%. Removing Stage B leads to an average L2 of 0.78m and a collision rate of 0.49%. The largest differences appear at longer horizons. The 3-second L2 decreases from 1.40-1.41m (without Stage A or without Stage B) to 1.25m when all stages are active, while the 1-second L2 remains close at 0.23m. This shows that the combination of Stage A and Stage B mainly strengthens long-horizon stability.

**(ii) With ego status.** With all stages active, KEPT reaches an average L2 error of 0.63m and a collision rate of 0.16%. Removing Stage A or Stage B increases the average L2 to 0.75m, with collision rates of 0.45% and 0.51%, respectively. Differences at shorter horizons remain small, while most of the improvements come from longer horizons. The 3-second L2 decreases from 1.39m or 1.36m to 1.13m, and the 2-second L2 decreases from 0.64m or 0.66m to 0.56m when combining all three stages together.

Across both usage modes, having Stage A and Stage B together on top of Stage C is crucial. Removing either stage weakens long-horizon accuracy and increases collision risk, while the full triple-stage fine-tuning strategy consistently ensures the most accurate and safest trajectories. This result is consistent with the design goals outlined in Section 3.4.

#### 4.3.2 Ablation Study on the Top-K Value of RAG

Table 4 evaluates how many retrieved examples ($K$) should be provided. All variants use the same triple-stage fine-tuning strategy so that the effect of retrieval depth could be examined in isolation. In line with the RAG pipeline in Section 3.3 and the CoT integration in Section 3.5, the results follow a concave pattern. A small Top-K offers useful priors that complement the target scene, while a large Top-K would weaken the planner's reasoning.

**(i) Without ego status.** Increasing the retrieval depth from Top-0 (without any retrieval scens) to Top-2 improves both accuracy and safety. The average L2 decreases from 0.73m to 0.70m, and the average collision rate decreases from 0.26% to 0.21%. The largest improvements appear at the 3-second horizon, where the L2 drops from 1.30m to 1.23m and the collision rate decreases from 0.55% to 0.44%. When $K$ is increased beyond 2, the trend reverses, which suggests that to many retrieved scenes introduce noise

| **Top-K Retrieval** | **L2 (m)↓** | | | | **Collision (%)↓** | | | |
|---|---|---|---|---|---|---|---|---|
| | 1s | 2s | 3s | Avg. | 1s | 2s | 3s | Avg. |
| Top-0 (w/o RAG) | 0.24 | 0.66 | 1.30 | 0.73 | 0.06 | 0.18 | 0.55 | 0.26 |
| Top-1 | 0.23 | 0.64 | 1.25 | 0.71 | 0.06 | 0.16 | 0.51 | 0.24 |
| **Top-2** | **0.23** | **0.63** | **1.23** | **0.70** | **0.05** | **0.14** | **0.44** | **0.21** |
| Top-3 | 0.25 | 0.68 | 1.27 | 0.73 | 0.07 | 0.19 | 0.59 | 0.28 |
| Top-4 | 0.31 | 0.78 | 1.41 | 0.83 | 0.09 | 0.23 | 0.74 | 0.35 |
| Top-0* (w/o RAG) | 0.21 | 0.58 | 1.15 | 0.65 | 0.05 | 0.15 | 0.38 | 0.19 |
| Top-1* | 0.21 | 0.56 | 1.13 | 0.63 | 0.03 | 0.13 | 0.31 | 0.16 |
| **Top-2*** | **0.19** | **0.52** | **1.12** | **0.61** | **0.02** | **0.11** | **0.29** | **0.14** |
| Top-3* | 0.22 | 0.58 | 1.17 | 0.66 | 0.07 | 0.19 | 0.44 | 0.23 |
| Top-4* | 0.24 | 0.62 | 1.26 | 0.71 | 0.08 | 0.21 | 0.53 | 0.27 |

**Table 4** Ablation study on the Top-K value of RAG. ↓ indicates lower is better. $*$ indicates the use of ego status as inputs. Only the NoAvg protocol is adopted. ↓ indicates lower is better. The VLM backbone is Qwen2-VL-2B.

rather than helpful context.

**(ii) With ego status.** The Top-2* configuration reaches an average L2 of 0.61m and an average collision rate of 0.14%. This represents an improvement over Top-0*. Long-horizon metrics are also improved, with the 3-second L2 decreasing from 1.15m to 1.12m and the collision rate decreasing from 0.38% to 0.29%. As before, using a large Top-K value leads to worse performance.

A small but effective retrieval set works best in our pipeline. Using $K = 2$ consistently provides the strongest balance between accuracy and safety. This matches the motivation in Section 3.3 and Section 3.5. For this reason, we use Top-2 as the default setting for main results in Section 4.2.

### 4.3.3 Ablation Study on Different Retrieval Strategies

To evaluate the effectiveness of our retrieval strategy, we conduct an ablation study comparing the proposed k-means & HNSW pipeline with a simple similarity search, a k-means only variant, and a standard HNSW search. Since clustering and HNSW indexing have little influence on retrieval accuracy but significantly reduce retrieval time, we focus on their efficiency benefits. For this ablation, we report the average time required to retrieve results for a single test scene, which serves as the primary measure

| Retrieval Strategy | Top-1 Avg. | Top-2 Avg. | Top-3 Avg. | Top-4 Avg. | Top-5 Avg. | Overall Avg. |
|---|---|---|---|---|---|---|
| Simple Search | 0.766 | 0.783 | 0.748 | 0.796 | 0.756 | 0.770 |
| k-means only | 0.162 | 0.159 | 0.149 | 0.150 | 0.150 | 0.154 |
| HNSW only | 0.032 | 0.033 | 0.031 | 0.032 | 0.034 | 0.032 |
| **k-means & HNSW** | **0.014** | **0.014** | **0.015** | **0.013** | **0.014** | **0.014** |

**Table 5** Ablation study on the retrieval time (ms) of different retrieval strategies.

of retrieval performance.

For each retrieval strategy, we evaluate performance using Top-K values from 1 to 5. We then compute the final metric as the arithmetic mean across these Top-K settings, giving a balanced view of retrieval speed. As shown in Table 5, simple cosine-similarity search already achieves millisecond-level performance on the database of nearly ten thousand scene embeddings. Adding k-means clustering lowers the average query time to about 0.1ms. HNSW offers an additional improvement, reducing the retrieval latency to roughly 0.03ms. Our k-means & HNSW method achieves the best efficiency, with an average retrieval time of 0.014ms per query. All retrieval tests are run on a single Intel i7-13700 CPU without GPU acceleration.

#### 4.3.4 Ablation Study on Various Vision Encoders

| Vision Encoder | L2 (m)↓ | | | | Collision (%)↓ | | | |
|---|---|---|---|---|---|---|---|---|
| | 1s | 2s | 3s | Avg. | 1s | 2s | 3s | Avg. |
| Temporal Spatial-only* | 0.22 | 0.61 | 1.23 | 0.69 | 0.04 | 0.19 | 0.41 | 0.21 |
| Temporal Frequency-only* | 0.23 | 0.62 | 1.25 | 0.70 | 0.04 | 0.21 | 0.48 | 0.24 |
| **TFSF*** | **0.21** | **0.56** | **1.13** | **0.63** | **0.03** | **0.13** | **0.31** | **0.16** |

**Table 6** Ablation study on vision encoders. Only the NoAvg protocol is adopted. ↓ indicates lower is better. $*$ indicates the use of ego status as inputs. The Top-K value of RAG is set to 1 accordingly. The VLM backbone is Qwen2-VL-2B.

To validate the design of the TFSF encoder, we compare three vision encoders in this ablation, namely a temporal spatial-only variant that aggregates per-frame features over time in the image spatial domain, a temporal frequency-only variant that operates purely in the temporal frequency domain, and the full TFSF encoder which combines both branches. The results are summarized in Table 6.

The numbers show a consistent pattern. Removing either branch leads to a degradation in both L2 and collision rate, while the full TFSF encoder achieves the best performance across all horizons. The gap is particularly clear at 2-3 s, where single-branch encoders are more prone to lateral drift and delayed reactions.

The temporal spatial branch maintains a high-fidelity, multi-scale description of the scene layout as frames evolve, which is crucial for understanding static obstacles and road geometry. In contrast, the temporal frequency branch emphasizes temporal variation in the feature space, it helps the encoder remain stable when objects move in and out of view or when lighting conditions change suddenly. When only the spatial branch is used, the encoder lacks the robustness to blocks and inadequate lighting, and its long-horizon predictions become more imprecise. Respectively, when only the frequency branch is used, the vision encoding loses spatial precision. The full TFSF encoder brings these two perspectives together, therefore achieves better performance in downstream tasks.

#### 4.3.5 Ablation Study on Various VLM Backbones

| VLM Backbone | L2 (m)↓ | | | | Collision (%)↓ | | | | Avg. Inf. Time (s)↓ |
|---|---|---|---|---|---|---|---|---|---|
| | 1s | 2s | 3s | Avg. | 1s | 2s | 3s | Avg. | |
| Qwen2-VL-2B* | 0.21 | 0.56 | 1.13 | 0.63 | 0.03 | 0.13 | 0.31 | 0.16 | **7.36** |
| Qwen2-VL-7B* | <u>0.20</u> | <u>0.53</u> | <u>1.09</u> | <u>0.61</u> | 0.03 | <u>0.11</u> | <u>0.28</u> | <u>0.14</u> | 11.27 |
| Qwen2.5-VL-3B* | 0.21 | 0.54 | 1.11 | 0.62 | 0.03 | 0.13 | 0.33 | 0.16 | <u>8.11</u> |
| Qwen2.5-VL-7B* | **0.19** | **0.52** | **1.07** | **0.59** | 0.03 | **0.10** | **0.26** | **0.13** | 14.43 |

**Table 7** Ablation study on VLM backbones. Only the NoAvg protocol is adopted. ↓ indicates lower is better. $*$ indicates the use of ego status as inputs. The Top-K value of RAG is set to 1 accordingly.

To examine how KEPT performs with different VLM Backbones, we replace Qwen2-VL-2B in our pipeline with three other Qwen-series VLMs, namely Qwen2-VL-7B, Qwen2.5-VL-3B, and Qwen2.5-VL-7B. Table 7 summarizes the L2 displacement error, collision rate, and average inference time, which is crucial for potential application of the framework.

The results indicate that increasing the parameter scale of VLM backbones can indeed bring measurable benefits on both L2 and collision rate metrics. Both 7B VLMs improve upon the 2B baseline, and Qwen2.5-VL-7B achieves the lowest errors, with consistent gains at 1s, 2s, and 3s. Since the parameter scale of Qwen2.5-VL-3B is comparable to that of Qwen2-VL-2B, their overall performance is also quite similar.

At the same time, this ablation study also makes clear that these gains come with a substantial increase in inference time. Both 7B VLMs, although achieve better performance, still incur noticeable slow-downs. From the perspective of an autonomous driving system, a longer inference time directly increases computational cost and could significantly harm real-time performance. In addition, the 7B VLMs are clearly harder to deploy on a vehicle than the 2B VLM, especially once RAG is enabled and the prompt becomes longer, which further raises the memory requirements.

In summary, this ablation shows that KEPT could benefit from larger VLMs. However, when jointly considering accuracy and computational cost, Qwen2-VL-2B offers the most balanced trade-off. We therefore adopt Qwen2-VL-2B as the default VLM backbone for all main experiments, and regard the 7B VLMs as upper-bound variants that illustrate the potential of the frameworkd when latency constraints are relaxed.

### 4.4 Case Studies

In this subsection, we examine six representative scenes from the nuScenes official test set: three long-tail success and three failure cases to evaluate model behavior in challenging, long-tail conditions. For the failure cases, we also demonstrate how prompt refinement could benefit the predicted trajectories.

#### 4.4.1 Long-Tail Success Case Analysis

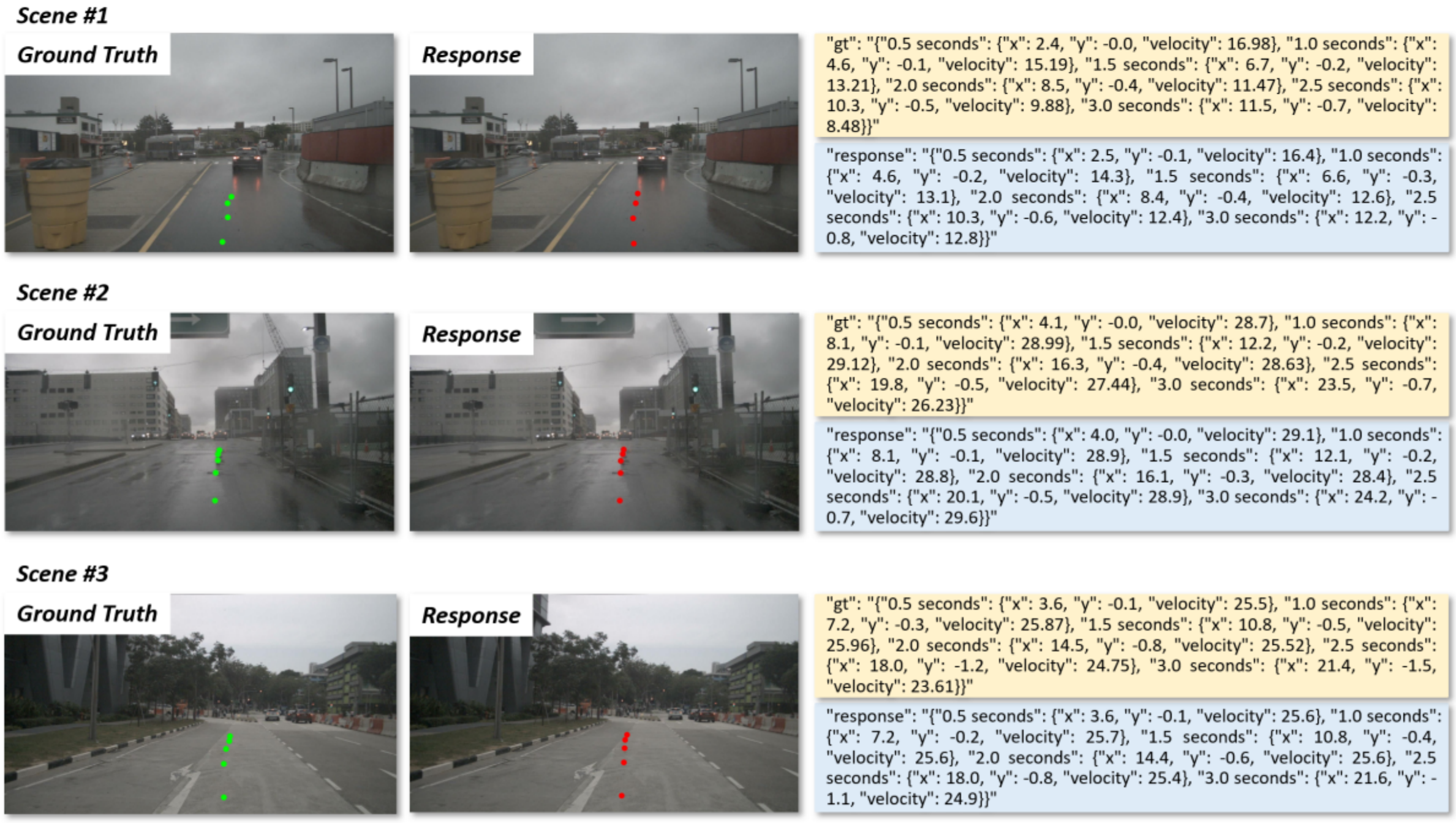

**Fig. 6** Long-tail successes. Green denotes the ground truth trajectory, and red denotes the model response. Some trajectory way-points cannot be fully shown in the front-view image due to field-of-view limits. The Top-K value of RAG is set to 2 accordingly. The VLM backbone is Qwen2-VL-2B.

In Fig. 6, we demonstrate how KEPT provides feasible trajectories in three long-tail cases. Scene #1 shows the ego vehicle is entering an non-standard intersection in relatively low visibility. Temporary barriers, the bus, and weak lane markings make it difficult to determine the correct path. Scene #2 shows the ego vehicle is driving along an urban straight road in heavy rain. The wet surface produces strong reflections and the right side is occupied by a construction area. Scene #3 shows the ego vehicle is approaching a non-standard intersection with ambiguous lane boundaries. In all demonstrated cases, KEPT could predict trajectories with good alignment to ground truth trajectories.

These successful cases provide a concrete view of how KEPT performs in such long-tail situations. The TFSF encoder helps to capture both frequency and spatial patterns when parts of the scene are occluded or when illumination varies from frame to frame. The retrieved reference scenes provide additional prior knowledge, which are further combined with CoT prompts to ensure feasible trajectories. These components together guarantee reasonable motion planning even when the road layout and visual condition is not ideal.

#### 4.4.2 Long-Tail Failure Case Analysis

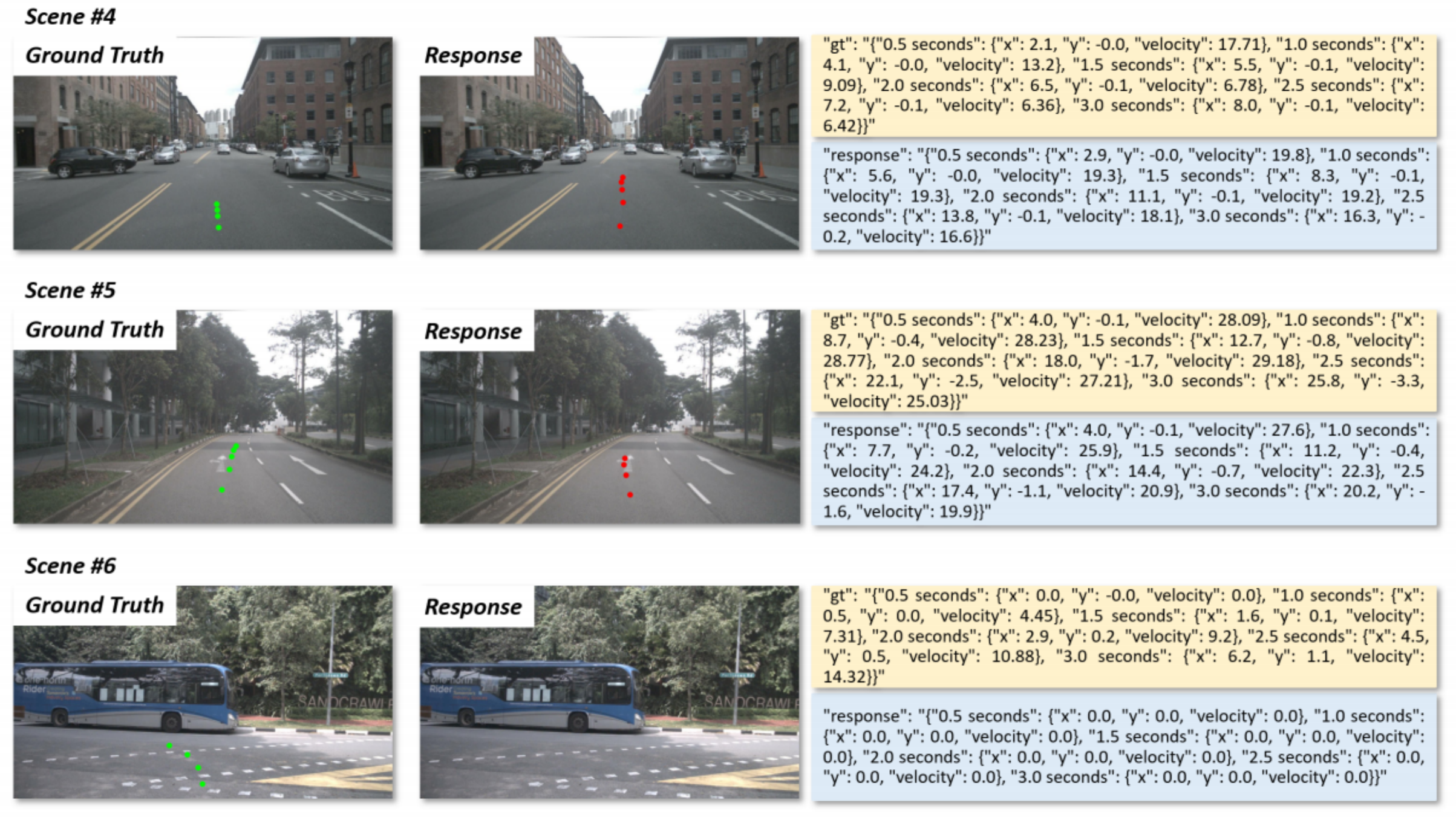

**Fig. 7** Failure case analysis. Green denotes the ground truth trajectory, and red denotes the model response. Some trajectory way-points cannot be fully shown in the front-view image due to field-of-view limits. The Top-K value of RAG is set to 2 accordingly. The VLM backbone is Qwen2-VL-2B.

In Fig. 7, we also demonstrate that KEPT might fail in scene understanding and trajectory predicting in several cases. Scene #4 shows the ego vehicle is approaching a T-junction with a car trying to merge from the left side road. Scene #5 shows the ego vehicle is driving along an urban road with the current lane occupied by a slow vehicle ahead. Scene #6 shows the ego vehicle is trying to turn left at an irregular intersection with a bus stopped ahead.

In Scene #4, the ego vehicle is expected to decelerate to avoid collision with the merging car as the ground truth trajectory shows. However, the predicted trajectory by KEPT does not tend to decelerate, which might lead to potential collision. In Scene #5, the ego vehicle might change lane to the right early to ensure safety, while the predicted trajectory seems to hold the current lane. In Scene #6, the ground truth trajectory indicates that it is safe for the ego vehicle to turn left, but KEPT chooses to stop at the intersection, as all predicted way-points is $(0.0, 0.0)$.

Since the overall retrieval and fine-tuning paradigm is fixed, possible improvement of the predicted trajectories is expected with slight modifications on prompts. We therefore refine the CoT prompts according to different failure cases. Specifically, we change the listed *considerations* in the CoT prompting paradigm in Fig. 5 according to different failure cases, and the refined parts are listed in Fig. 8.

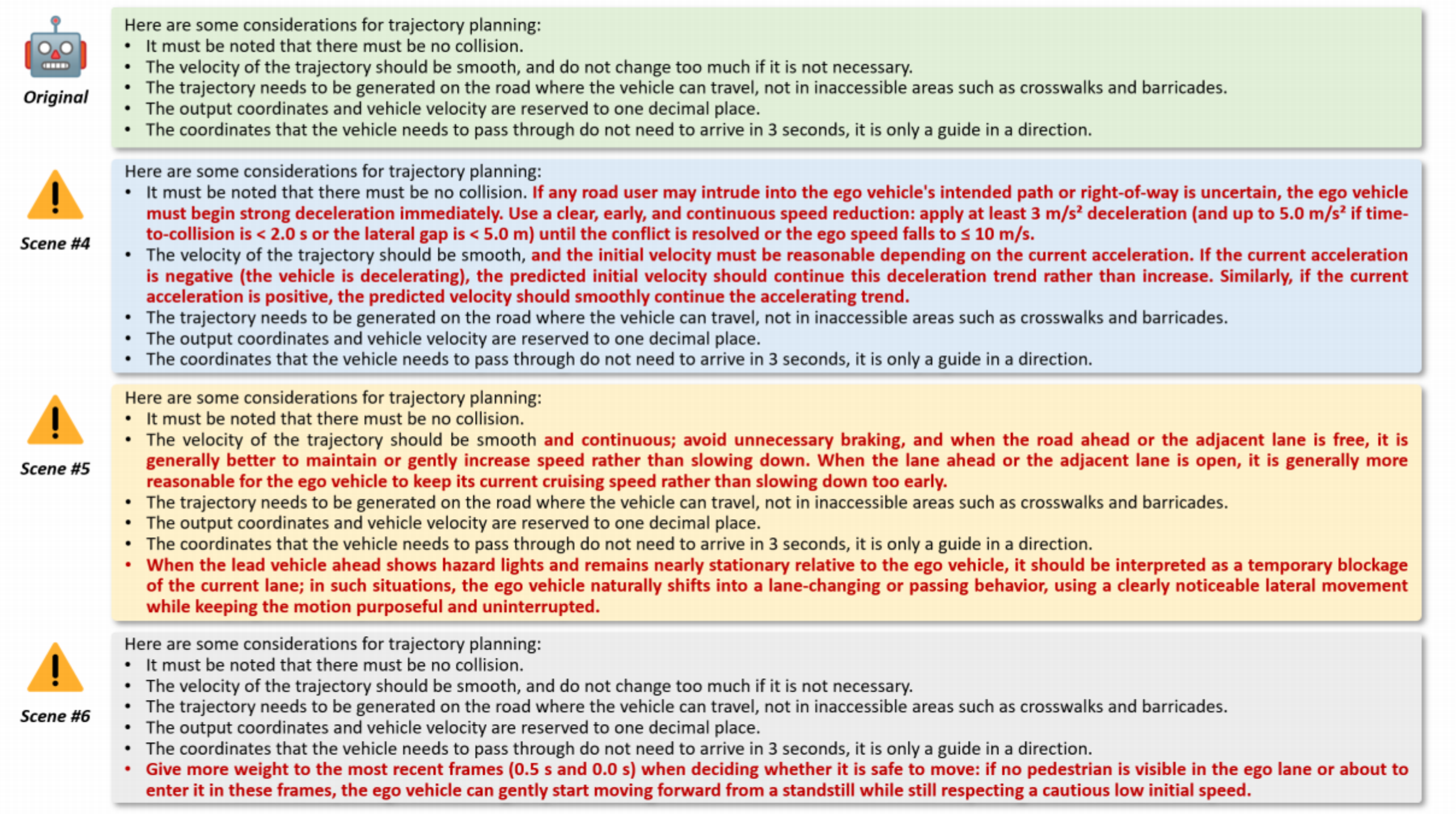

**Fig. 8** The refined prompts according to different failure cases. Only the *considerations* part of system prompts is modified. The supplementary considerations aims to guide the VLM to understand the current scene better, and to execute more appropriate actions.

In Fig. 9, we demonstrate the improved trajectories by refining prompts of the failure cases. In Scene #4, with the help of the modified prompts, KEPT captures the merging tendency of the car in the intersection and therefore generates a trajectory of slowing down. In Scene #5, KEPT notices the slow car ahead and recognizes the potential blockage of the current lane, and therefore decides to change lane to the right. In Scene #6, according to the refined prompt, KEPT regards the current situation as safe and therefore generates a left-turn trajectory. All improved trajectories keep good alignment with the ground truth, suggesting that the qualify of generated trajectories could benefit from modifications on the CoT prompts in case of failure.

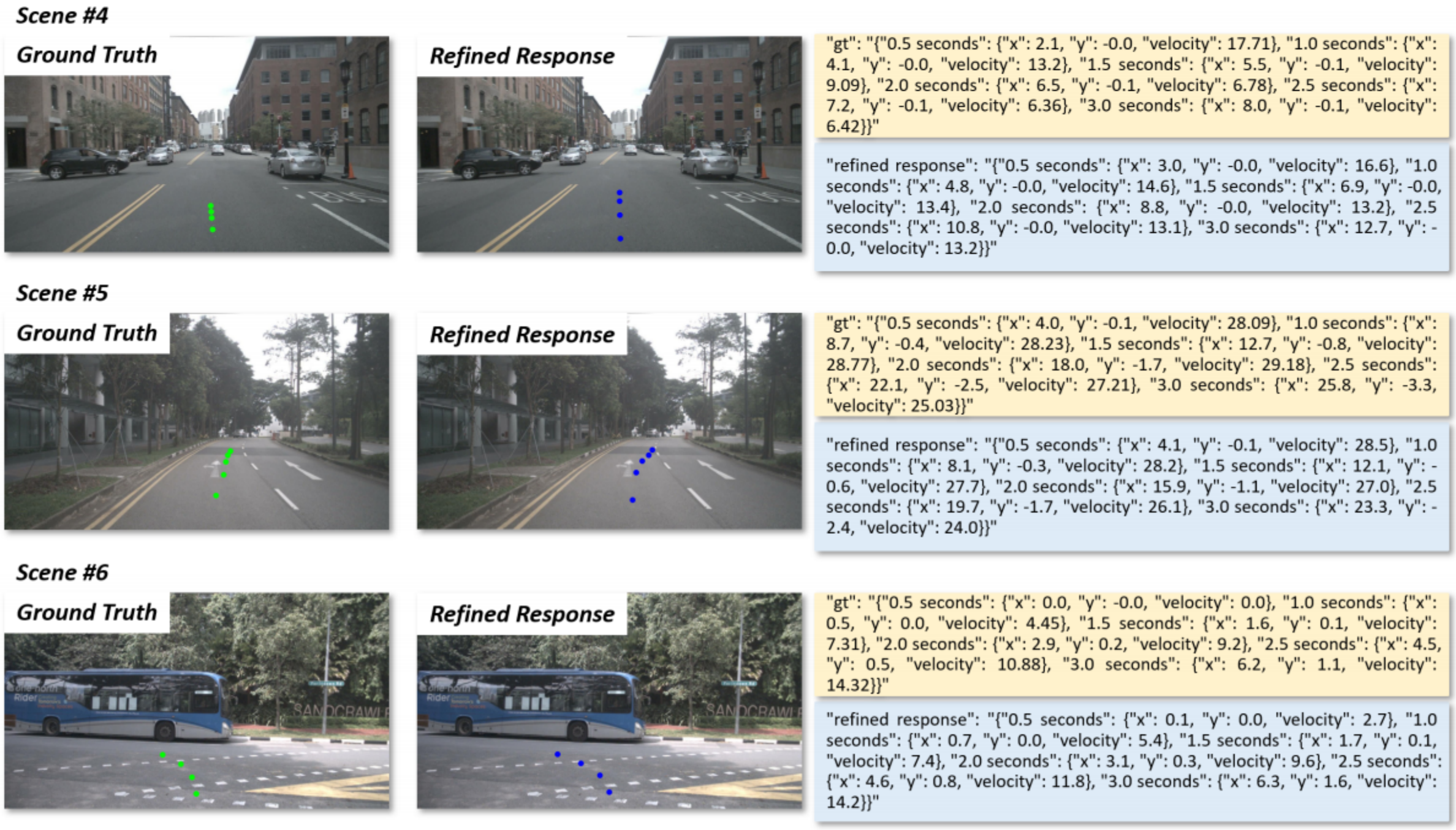

**Fig. 9** Improved trajectories by refining prompts of the failure cases. Green denotes the ground truth trajectory, and red denotes the model response. Some trajectory way-points cannot be fully shown in the front-view image due to field-of-view limits. The Top-K value of RAG is set to 2 accordingly. The VLM backbone is Qwen2-VL-2B.

# 5 Conclusion

In this paper, we introduce KEPT (Knowledge-Enhanced Prediction of Trajectories), a retrieval-augmented VLM framework for short-horizon trajectory prediction from consecutive driving frames. KEPT integrates a TFSF encoder that produces motion-aware scene embeddings, a scalable retrieval module based on k-means clustering and HNSW indexing, a triple-stage fine-tuning strategy for the VLM backbone, and a CoT prompting paradigm designed around planning constraints.

Experiments on the nuScenes dataset show that KEPT improves a range of baseline methods across both open-loop protocols. Under NoAvg, it achieves lower L2 errors and collision rates than recent modular and VLM-based planners. Under TemAvg, it stays competitive without ego-status inputs and delivers the best accuracy-safety balance when ego information is provided. Ablation studies highlight two key factors. Stages A, B, and C should work together, for removing either spatial perception or trajectory-supervised reasoning fine-tuning harms long-horizon accuracy and safety. Also, Top-2 retrieval consistently offers the clearest gains by providing useful priors without adding noise. Finally, the proposed k-means & HNSW retrieval achieves sub-millisecond latency, supporting real-time use.

Despite the gains, KEPT has several limitations. Our retrieval priors are built from nuScenes-like distributions; coverage and clustering granularity bound performance, and long-tail or out-of-domain scenes (e.g., extreme weather, unusual road rules) may yield suboptimal priors. The TFSF encoder currently uses seven 2 Hz front-view frames and predicts short horizons; richer sensors and longer temporal context could improve stability. CoT prompting adds variability and modest latency, and our constraint set is partly hand-crafted. Finally, we report open-loop metrics; closed-loop, real-time evaluations on a vehicle are required to verify safety and robustness.

Future work will therefore (i) extend KEPT to multi-sensor inputs and cross-city generalization with domain adaptation, (ii) develop active and safety-aware retrieval that updates the knowledge base in real time. In addition, RL-based alignment (e.g., GRPO) is orthogonal and complementary to our present approach. A principled integration would couple our supervised LoRA stages with (a) a reward model encoding multi-objective driving preferences (e.g., comfort, legality, social compliance) and (b) simulation- or dataset-in-the-loop rollouts for closed-loop credit alignment. We anticipate two promising paths: The first one is LoRA followed with GRPO, where LoRA provides stable metric grounding and GRPO performs light post-training for preference trade-offs; the second one is constrained RL, where retrieval-supplied priors and CoT constraints act as safety shields during policy updates. We leave these preference-centric extensions to future work, as they require additional infrastructure (reward definition, rollouts, and safety monitors) beyond the open-loop, metric-supervised scope of this paper. We believe these directions, together with the lightweight nature of the Qwen-series backbones used here, make KEPT a promising foundation for trustworthy, interpretable, and scalable VLM planning in autonomous driving.

## Replication and Data Sharing

The codes are available at https://github.com/yjwangtj/KEPT.

The JSONs of datasets used are available at https://huggingface.co/datasets/larswang/kept_datasets/tree/main.

## Declaration of Competing Interest

The authors declare that they have no known competing financial interests or personal relationships that could have appeared to influence the work reported in this paper.

## Acknowledgment

This research was supported by National Key R&D Program of China (2023YFB2504400), the National Nature Science Foundation of China (No. 62088101, No. 62373289, No. 62273256 and No. 62473291), Shanghai Municipal Science and Shanghai Automotive Industry Science and Technology Development Foundation (No.2407), and the Fundamental Research Funds for the Central Universities.